\documentclass{article}

     \usepackage[preprint,nonatbib]{neurips_2023}

\usepackage[square,numbers]{natbib}
\usepackage[table]{xcolor}
\usepackage{rotating}
\usepackage[flushleft]{threeparttable}
\usepackage[utf8]{inputenc} % allow utf-8 input
\usepackage[T1]{fontenc}    % use 8-bit T1 fonts
\usepackage{url}            % simple URL typesetting
\usepackage{booktabs}       % professional-quality tables
\usepackage{amsfonts}       % blackboard math symbols
\usepackage{nicefrac}       % compact symbols for 1/2, etc.
\usepackage{microtype}      % microtypography
\usepackage{xcolor}         % colors
\usepackage{enumitem} % added <<<<<<<<<<<<<
\usepackage{algorithm2e}% added <<<<<<<<<<<<<<
\definecolor{mydarkblue}{rgb}{0,0.08,0.45}
\usepackage[colorlinks,citecolor=mydarkblue,urlcolor=mydarkblue,linkcolor=mydarkblue,filecolor=mydarkblue]{hyperref}
\usepackage{multirow}
\usepackage{graphicx}  
\usepackage{makecell}
\usepackage{wrapfig}
\usepackage{amsmath,amsfonts,bm}

\def\eqref#1{equation~\ref{#1}}
\def\1{\bm{1}}

\def\vx{{\bm{x}}}

\def\vz{{\bm{z}}}

\DeclareMathAlphabet{\mathsfit}{\encodingdefault}{\sfdefault}{m}{sl}
\SetMathAlphabet{\mathsfit}{bold}{\encodingdefault}{\sfdefault}{bx}{n}

\title{Building One-class Detector for \textit{Anything}: Open-vocabulary Zero-shot OOD Detection Using Text-image Models}

\author{%
\textbf{Yunhao Ge$^{1,3*}$,  Jie Ren$^{2*}$, Jiaping Zhao$^{1}$, Kaifeng Chen$^{1}$, Andrew Gallagher$^{1}$,  Laurent Itti$^{3}$,} \\ \textbf{Balaji Lakshminarayanan$^{2}$}  \\
$^{1}$Google Research \ \ $^{2}$Google DeepMind \ \  $^{3}$University of Southern California\\  
{\tt \scalebox{1}{
$^{*}$co-first authors \ \ \ Email:\{yunhaoge@usc.edu, jjren@google.com\}}}
}

  \usepackage[compact]{titlesec}
  \titlespacing{\section}{0pt}{1ex}{0ex}
  \titlespacing{\subsection}{0pt}{1ex}{0ex}
  \titlespacing{\subsubsection}{0pt}{0.5ex}{0ex}

\begin{document}

\maketitle
\vspace{-1em}

\begin{abstract}
We focus on the challenge of out-of-distribution (OOD) detection in deep learning models, a crucial aspect in ensuring reliability. Despite considerable effort, the problem remains significantly challenging in deep learning models due to their propensity to output over-confident predictions for OOD inputs. We propose a novel one-class open-set OOD detector that leverages text-image pre-trained models in a zero-shot fashion and incorporates various descriptions of in-domain and OOD. Our approach\footnote{Work carried out mainly at Google} is designed to detect anything not in-domain and offers the flexibility to detect a wide variety of OOD, defined via fine- or coarse-grained labels, or even in natural language. We evaluate our approach on challenging benchmarks including large-scale datasets containing fine-grained, semantically similar classes, distributionally shifted images, and multi-object images containing a mixture of in-domain and OOD objects. Our method shows superior performance over previous methods on all benchmarks.\\
Code is available at \small{\textcolor{magenta}{\url{https://github.com/gyhandy/One-Class-Anything}}}

%   The abstract paragraph should be indented \nicefrac{1}{2}~inch (3~picas) on
%   both the left- and right-hand margins. Use 10~point type, with a vertical
%   spacing (leading) of 11~points.  The word \textbf{Abstract} must be centered,
%   bold, and in point size 12. Two line spaces precede the abstract. The abstract
%   must be limited to one paragraph.
\end{abstract}

\section{Introduction}

Out-of-distribution detection (OOD) is essential for ensuring the reliability of machine learning systems. 
When a machine learning system is deployed in the real world, it may encounter unexpected abnormal inputs that are not from the same distribution as the training data. 
Detection and removal of OOD inputs prevents the machine learning (ML) system from making incorrect predictions that could otherwise lead to serious failures especially in safety-critical applications.
For example, a person classification model is trained to localize people in images. If an image that does not contain a person, but instead contains an animal or a sculpture, the model may erroneously label the non-person as a person.
%In life-critical applications such as autonomous driving systems, correctly identifying persons is of paramount importance.
Accurate and reliable one-class detection is of paramount importance in life-critical applications; for example, correctly perceiving persons in autonomous driving systems.

% skin diseases diagnosis, the system should be able to identify and remove images that do not contain any skin lesions such as an image of cloth or a necklace. 
% Similarly in 
 %the system should be able to detect images that do not contain any traffic signs such as billboards and mailboxes so that the model will not falsely predict them as a traffic sign.

Although OOD detection has been studied previously in traditional ML models \cite{scholkopf2000support,emmott2015meta}, deep learning models are known to output over-confident predictions for OOD inputs, making OOD detection in deep learning much more challenging. 
Recent efforts have focused on developing methods to correct for the naive softmax probability \cite{hendrycks2016baseline,liang2017enhancing,hendrycks2019scaling,liu2020energy}, using deep neural representations to measure the distance to the training distribution \cite{lee2018simple,ren2021simple,sun2022out,fort2021exploring}, leveraging OOD data to learn a more precise decision boundary between in-domain and OOD \cite{hendrycks2018deep,roy2022does}, and using deep density models to measure the likelihood under the training distribution \cite{choi2018waic,ren2019likelihood,nalisnick2018deep,morningstar2021density}.
% (3) contrastive learning based methods which incorporatethe contrastive loss into the classification cross-entropy loss to improve representation learning and consequently improve OOD detection (Winkens et al., 2020; Zhou et al., 2021). 

\begin{figure}[ht]
\vspace{-20pt}
\centering
\includegraphics[width=0.9\textwidth]{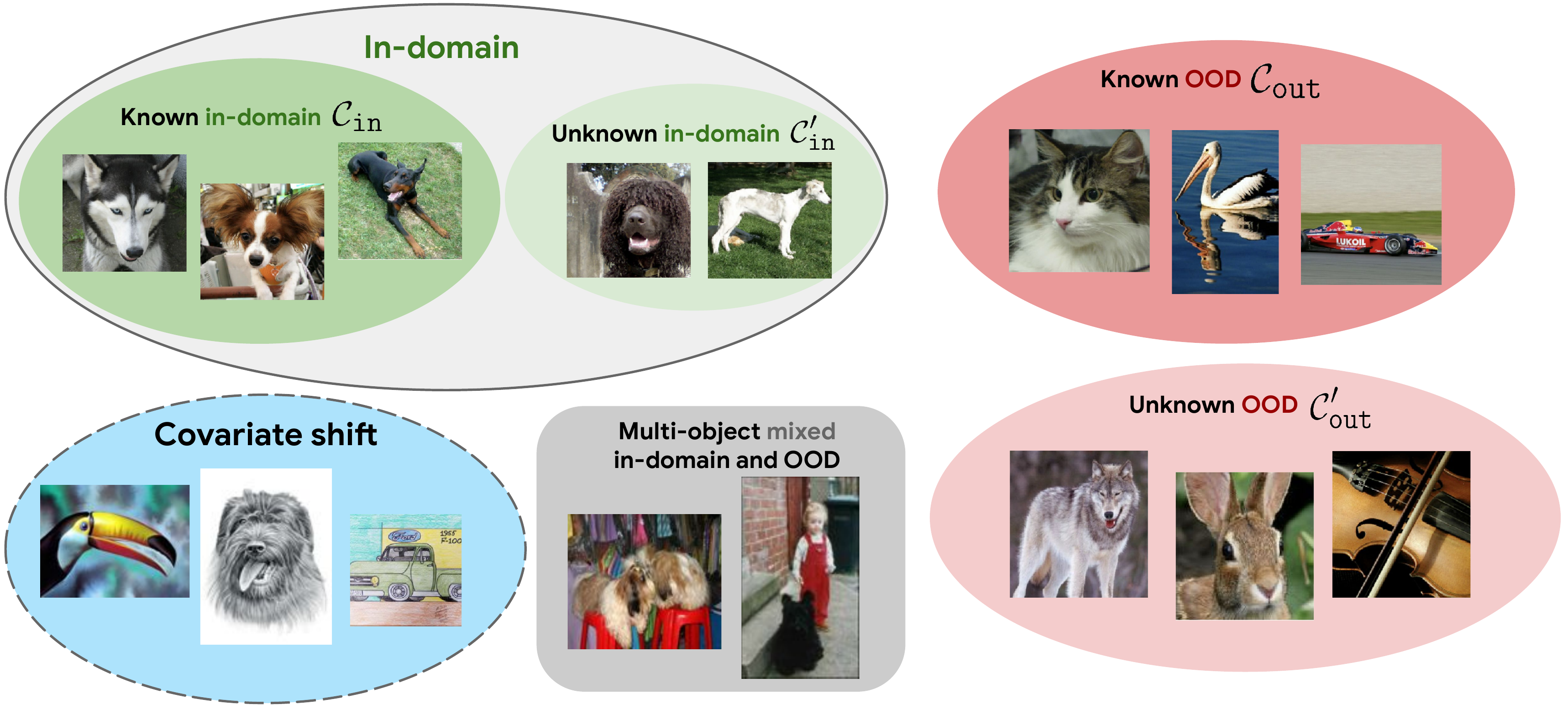}
\vspace{-0.5em}
\caption{
We study the one-class OOD detection problem where OOD can be anything not in-domain. 
Example of building a dog detector with some known dogs  $\mathcal{C}_{\texttt{in}}=$\{Husky, Papillon, Dobermann\} and known non-dogs $\mathcal{C}_{\texttt{out}}=$\{Cat, Bird, Person\}.  When deploying such a one-class detector in the real world, it's important for it to be robust to several different types of shifts: (1)
Unknown in-domain classes include new species of dogs, and unknown OOD such as wolves, bunnies, violins, (2) Multi-object cases (cats along with dogs, persons along with dogs), (3) Covariate shift (drawings of dogs, painting of a bird, cartoon car, etc).
% We study the one-class OOD detection problem where OOD can be anything not in-domain. We consider not only the detection of known in-domain and OOD, but also the unknown in-domain, unknown OOD, distribution shifted, and multi-label settings, to mimic the real world scenarios.
}
\vspace{-0.5em}
\label{fig:overall}
\end{figure}

For evaluation, most OOD detection methods use one dataset as in-domain and another dataset as OOD. For example, the CIFAR\-100 vs CIFAR\-10 \cite{ren2019likelihood,ren2021simple,fort2021exploring,lee2018simple}, or ImageNet vs Places365 \cite{hendrycks2019scaling,ming2022delving} benchmark datasets; both have in-domain data that consists of a set of classes, and OOD data comprised of a different set of classes that are non-overlapping with the in-domain.
That scenario treats OOD classes as a closed set. 
However, in realistic scenarios, in-domain data often consists of a set of classes belonging to a unified high-level superclass, and OOD data is \textit{anything} that is not in that superclass, which can be viewed as an \textit{open-set one-class} anomaly detection problem \cite{khan2010survey,noumir2012simple}. 
For example, the OOD detection in the person identification scanario is to detect anything that is non-person.
% and the OOD detection in the traffic sign classification model is to detect anything that has no traffic sign. 

The most straightforward approach for building a one-class OOD detector is to train a binary classifier for in-domain and OOD classes \cite{bitterwolfrevisiting}. However, since the OOD space is large (anything that is not in-domain) and often includes classes that follow a long-tail distribution \cite{roy2022does}, it is impossible to include all the possible OOD data when training. Consequently, a model trained on a subset of OOD data may suffer from poor generalization to the unseen OOD at test time. 
Another approach is to learn the distribution of the in-domain data without using prior knowledge of OOD \cite{ruff2018deep,chalapathy2018anomaly,zhang2021anomaly,hojjati2021dasvdd,park2021wrong,morningstar2021density,chalapathy2017robust}, but that approach cannot leverage the knowledge of known OOD data which can help with learning a more precise boundary. 
Another issue with the existing one-class OOD work is that the datasets used for evaluation are fairly simple and small scale as MNIST and SVHN. Usually one class serves as in-domain and the remaining classes are OOD (such as using 1 as in-domain and the numbers 2 through 9 as OOD). 
That testing regime does not fully explore a setting with the open-set assumption that OOD can be anything not in-domain, including, for example, images of non-numbers.

Recently, large pretrained text-image models such as CLIP (Contrastive Language-Image Pretraining) \cite{radford2021learning}, and LiT \cite{zhai2022lit} learn the image and text representations simultaneously from massive image captioning data, and can be used as a zero-shot classifier at inference time by comparing the similarity between the class label and the image in the embedding space.
% The predicted class is the class label that has the highest similarity among all the candidate labels to the input image in the embedding space. 
% The zero-shot classifier is not only a powerful classifier. 
It has also shown that the maximum softmax score from the zero-shot classifier is a reasonably good confidence score for OOD detection \cite{ming2022delving}. 
However, that method only takes the in-domain class labels without leveraging the OOD information.
\citet{fort2021exploring,esmaeilpour2022zero} 
%Recent work \cite{fort2021exploring,esmaeilpour2022zero} 
propose \textit{a weaker form of outlier exposure (OE)} by including OOD class names into the label set without any accompanying images, and use the sum of the softmax over the OOD class labels as the OOD score. 
Since the methods were only evaluated on closed-set tasks, it is unclear how well the methods would perform on one-class OOD detection problem which evaluate the generalization ability with unseen classes.
We are also interested in evaluating on large-scaled datasets such as ImageNet, and its variants to see how well the methods perform on fine-grained and semantically similar classes, and on distributional shifted data.

In this work, we develop a one-class open-set OOD detector using text-image pretrained models in a zero-shot fashion. 
Our one-class open-set OOD detector detects \textit{anything} that is not in-domain, contrary to methods that specify a restricted set of predefined OOD classes. Our method can be used to detect \textit{any} type of OOD, defined either in fine-grained or coarse-grained labels, or even in natural language, through customizing the text labels in the text-image zero-shot classifier.  
We evaluate our method on large-scaled challenging benchmarks to mimic real-world scenarios, and test with
% (1) large-scaled dataset which contains fine-grained and semantically similar classes, 
(1) images from unseen classes, (2) distribution shifted images, (3) multi-object images that are may contain a mixture of in-domain and OOD objects. See Figure \ref{fig:overall} for an overview.
We show our method consistently outperforms the previous methods on all the challenging benchmarks. 
% Note that we focus on the outlier exposure setup because it is very easy to come up with a few OOD labels, even though the OOD images may be hard to collect. For example, we can use the labels ``animals'' and ``cars'' for non-person anomalies, and `mailbox' and `billboard' for non-traffic-sign anomalies.

Our contributions are the following: 
\begin{itemize} [leftmargin=1.5em,itemsep=-0.em]
    \item We find that  previous methods \cite{fort2021exploring,ming2022delving} do not work well on OOD detection for samples outside of the predefined class sets. 
    We propose better OOD scores that utilize the in-domain and OOD labels and show that they consistently perform best for detecting hard samples from long-tail unseen classes and under distribution shifts. 
    \item Our proposed method is flexible enough to incorporate various definition of in-domain and OOD. Because our method is based on text-image models, users can easily customize the definitions of in-domain and OOD via text labels, for example using class names at different hierarchical levels, or even including natural language sentences.  
    \item We tackle the challenging OOD detection for multi-object images that contain a mixture of in-domain and OOD objects. Integrating our scores into powerful segmentation models \cite{kirillov2023segment,liu2023grounding}, we are able to identify images with mixed in-domain and OOD objects, outperforming the baselines. 
    % Instead of evaluating the methods on relatively simple datasets as previous work does, we evaluate the methods on real and challenging scenarios. 
    % In particular, we evaluate datasets that contain: (1) fine-grained and semantically similar classes, (2) images that are distributional shift, and (3) 
    
\end{itemize}

\section{Methods}

\subsection{Background}

Contrastively pre-trained text-image models can be used as zero-shot classification models \cite{radford2021learning}. Text-image models consist of an image encoder $f_{\texttt{img}}(\cdot)$ and a text encoder $f_{\texttt{txt}}(\cdot)$. Given an input image $\vx$ and an input text $t$, the encoders produce embeddings $\vz_\texttt{img}=f_{\texttt{img}}(\vx)$ and $\vz_\texttt{txt}=f_{\texttt{txt}}(t)$ for the image and the text respectively. 
The model is trained to maximize the cosine similarity between the $\vz_\texttt{img}$ and $\vz_\texttt{txt}$ from the paired $\{\texttt{image}, \texttt{caption}\}$ data, and minimize the cosine similarity between the unpaired data.
At test time, to predict the class of an image $\vx$, we first encode the candidate class names $\mathcal{C}=\{c_1, \dots, c_n \}$ 
using the text encoder individually, $\{ \vz_\texttt{txt}^1, \dots, \vz_\texttt{txt}^n \}$. Then we compute the cosine similarity between the image $\vx$ and a set of candidate class names $\mathcal{C}$,
\begin{small}
\begin{align*}
\mathtt{logits}(\vx, \mathcal{C}) = \left[ \vz_\texttt{img} \cdot \vz_\texttt{txt}^1, \cdots, \vz_\texttt{img} \cdot \vz_\texttt{txt}^n  \right], 
\end{align*}
\end{small}
The predicted class for this image $\vx$ is $\hat{c}(\vx) = \mathrm{argmax}_c\ \mathtt{logits}(\vx, \mathcal{C})$.

For the problem of OOD detection, given a set of in-domain class labels $\mathcal{C}_{\texttt{in}} = \{c^{\texttt{in}}_1, \dots, c^{\texttt{in}}_N\}$, \citet{ming2022delving} propose to use the $\max_{c} p(c | \vx, \mathcal{C}_{\texttt{in}})$ as the confidence score, where $p(c| \vx, \mathcal{C}_{\texttt{in}})$ is the element of $\mathrm{softmax}\left(\mathtt{logits}(\vx,\mathcal{C}_{\texttt{in}})\right)$ corresponding to the label $c$, i.e. $p(c|\vx, \mathcal{C}_{\texttt{in}}) = \frac{e^{w_c}}{\sum_{j\in \mathcal{C}_{\texttt{in}}} e^{w_j}}, w_c = \vz_\texttt{img} \cdot \vz_\texttt{txt}^c$ . 
The $\mathtt{logits}$ can be further scaled by a temperature factor. 
A high confidence score indicates that the input image is likely to be from one of the $\mathcal{C}_{\texttt{in}}$ classes, and thus in-domain.
The corresponding OOD score is defined as
\begin{align}
\label{eq:max_prob_no_oe}
    % S_{\texttt{max\_prob\_without\_oe}} 
    S_{\texttt{-max\_prob}}(\vx) = - \max_{c\in\mathcal{C_{\texttt{in}}}} p(c | \vx, \mathcal{C}_{\texttt{in}})
    % \in\mathcal{C}_{\texttt{in}}
    % \in\mathcal{C}_{\texttt{in}}
\end{align}

The above method assumes that only a set of in-domain class labels is available, i.e., without exposure to OOD labels. However, although it can be difficult to obtain OOD images, it is often very easy to produce a set of possible OOD labels. In that setting, \cite{fort2021exploring,esmaeilpour2022zero} proposes to include the OOD labels $\mathcal{C}_{\texttt{out}}=\{c_1^{\texttt{out}}, \dots, c_M^{\texttt{out}}\}$ into the candidate label set $\mathcal{C} = \mathcal{C}_{\texttt{in}} \cup \mathcal{C}_{\texttt{out}}$, to utilize the knowledge as a weak form of outlier exposure, without using any OOD images for training. 
Then the $\mathtt{logits}$ are: 
% \begin{small}
\begin{align*}
\mathtt{logits}(\vx, \mathcal{C}_{\texttt{in}} \cup \mathcal{C}_{\texttt{out}}) =  \left[ \vz_\texttt{img} \cdot \vz_\texttt{txt}^1, \cdots, \vz_\texttt{img} \cdot \vz_\texttt{txt}^N, \vz_\texttt{img} \cdot \vz_\texttt{txt}^{N+1}, \cdots, \vz_\texttt{img} \cdot \vz_\texttt{txt}^{N+M}  \right] 
\end{align*}
% \end{small}
And the OOD score is defined as 
\begin{align}
\label{eq:sum_prob}
% S_{\texttt{sum\_prob}}(\vx) 
S_{\texttt{sum\_out\_prob}}(\vx)= \sum_{c\in\mathcal{C_{\texttt{out}}}} p(c| \vx, \mathcal{C}_{\texttt{in}} \cup \mathcal{C}_{\texttt{out}}),
% ; \mathcal{C}_{\texttt{in}} \cup \mathcal{C}_{\texttt{out}}
\end{align}
where $p(c| \vx, \mathcal{C}_{\texttt{in}} \cup \mathcal{C}_{\texttt{out}}) = \frac{e^{w_c}}{\sum_{j\in\mathcal{C_{\texttt{in}}}} e^{w_j} + \sum_{k\in\mathcal{C_{\texttt{out}}}} e^{w_k}}$,
$w_c =  \vz_\texttt{img} \cdot \vz_\texttt{txt}^c$.

\begin{figure}[t]
\vspace{-20pt}
\centering
\includegraphics[width=0.95\textwidth]{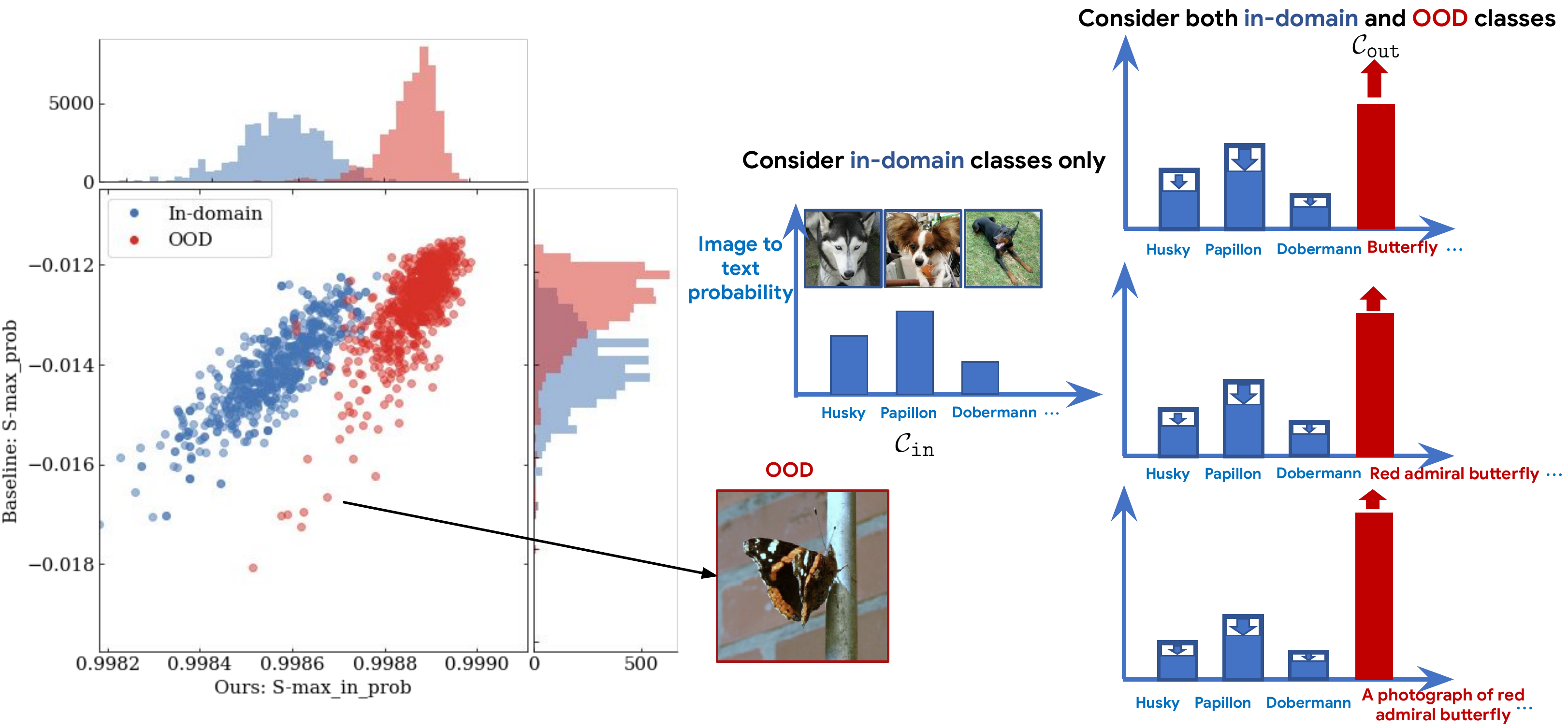}
\caption{Our methods utilize in-domain and OOD label sets. When in-domain classes $\mathcal{C}_{\texttt{in}}$ are comprised only of dog breeds, a butterfly may be mistaken for a Papillon dog, possibly due to the similar shape and color to the dog's ears. However, when a OOD set $\mathcal{C}_{\texttt{out}}$ is included consisting of the class name ``butterfly'', or a more precise butterfly type ``red admiral butterfly'', or a text description ``a photograph of red admiral butterfly'', the image embedding's similarity with this label pushes down the probabilities with the in-domain dog breeds, correctly identifying the image as OOD. Thus our method $S_{\texttt{-max\_in\_prob}}$ has better separation between in-domain and OOD compared to the baseline $S_{\texttt{-max\_prob}}$, as shown on the left 2D histograms.}
\label{fig:method}
\end{figure}

\subsection{Our methods: OOD scores utilizing in-domain and OOD label sets}

In this work, we follow on the same setting as \citet{fort2021exploring} because in the real scenario it is generally easy to produce a set of OOD class labels. 
For example, for the problem of detecting non-persons in a person detection system, it is easy to create a list of non-person labels that are commonly shown in photos, $\mathcal{C}_{\texttt{out}}=$\{animals, buildings, cars, food, $\dots$\}.
We would like to exploit these labels to improve the decision boundary between in- and out-of distribution. 
Note that $\mathcal{C}_{\texttt{out}}$ may not cover all the sub-types in OOD space due to the long-tail distribution. That is why we assume there are unseen classes $\mathcal{C}_{\texttt{out}}^\prime$.
Note that our methods are only based on $\mathcal{C}_{\texttt{out}}$ not $\mathcal{C}_{\texttt{out}}^\prime$. 
We will show the good generalization of our methods on unseen classes in Section \ref{sec:results_compare_methods}.

Suppose we have an in-domain label set $\mathcal{C}_{\texttt{in}}$ and a OOD label set $\mathcal{C}_{\texttt{out}}$. 
Inspired by \cite{ming2022delving}, we first propose the maximum softmax probability over the $\mathcal{C}_{\texttt{in}}$ as the confidence score, and its negative as the OOD score, 
\begin{align}
\label{eq:max_ind_softmax_all}
% S_{\texttt{max\_prob\_in}}(\vx) 
S_{\texttt{-max\_in\_prob}}(\vx)
= - \max_{c\in\mathcal{C_{\texttt{in}}}} p(c| \vx, \mathcal{C}_{\texttt{in}} \cup \mathcal{C}_{\texttt{out}}).
% ; \mathcal{C}_{\texttt{in}} \cup \mathcal{C}_{\texttt{out}}
\end{align}
Note that our score is different from \citet{ming2022delving} in the sense that we apply the softmax normalization over all labels $\mathcal{C} = \mathcal{C}_{\texttt{in}} \cup \mathcal{C}_{\texttt{out}}$, while \citet{ming2022delving} only applies softmax on $\mathcal{C}_{\texttt{in}}$.
Including $\mathcal{C}_{\texttt{out}}$ in the label sets is important for OOD detection, as shown in Figure \ref{fig:method}.
An OOD image may have relatively high similarity to one of the $\mathcal{C}_{\texttt{in}}$ classes due to spurious features. Once $\mathcal{C}_{\texttt{out}}$ is included, the OOD image's similarity with the OOD labels pushes down the probabilities with the in-domain $\mathcal{C}_{\texttt{in}}$ classes, correctly identifying the image as OOD.

Alternatively, we have the maximum softmax probability over the $\mathcal{C}_{\texttt{out}}$ as another candidate OOD score, 
\begin{align}
\label{eq:max_ood_softmax_all}
% S_{\texttt{max\_prob\_out}}(\vx) 
S_{\texttt{max\_out\_prob}}(\vx)
= \max_{c\in\mathcal{C_{\texttt{out}}}} p(c| \vx, \mathcal{C}_{\texttt{in}} \cup \mathcal{C}_{\texttt{out}}).
% ; \mathcal{C}_{\texttt{in}} \cup \mathcal{C}_{\texttt{out}}
\end{align}

We also consider a score based on $\mathtt{logits}$ without softmax normalization, as \citet{hendrycks2019scaling} previously show that in the larger-scale and real-world settings, the un-normalized maximum logit outperforms the normalized maximum softmax probability for OOD detection in the single modal models. 
We propose the OOD score as, 
\begin{align}
\label{eq:max_min}
% S_{\texttt{max\_logit\_diff}}(\vx) =
S_{\texttt{max\_logit\_diff}}(\vx) =
% & - \left( \max_{c\in\mathcal{C_{\texttt{in}}}} \mathtt{logits}(c| \vx) - \max_{c\in\mathcal{C_{\texttt{out}}}} \mathtt{logits}(c| \vx) \right) \\=
% & \max_{c\in\mathcal{C_{\texttt{out}}}} \mathtt{logits}(c| \vx) - \max_{c\in\mathcal{C_{\texttt{in}}}} \mathtt{logits}(c| \vx).
& \max_{d\in\mathcal{C_{\texttt{out}}}} w_d - \max_{c\in\mathcal{C_{\texttt{in}}}} w_c.
% ; \mathcal{C}_{\texttt{in}} \cup \mathcal{C}_{\texttt{out}}
\end{align}
The score measures if a test image has a higher similarity to any of the classes in $\mathcal{C}_{\texttt{out}}$ \textit{in comparison to} the similarity to any of the classes in $\mathcal{C}_{\texttt{in}}$. 
Having a reference for the similarity to $\mathcal{C}_{\texttt{in}}$ is helpful for understanding if the similarity to $\mathcal{C}_{\texttt{out}}$ is truly high or not. 
$S_{\texttt{max\_logit\_diff}}(\vx)>0$ suggests the image is more similar to classes in $\mathcal{C}_{\texttt{out}}$, and thus it can be inferred to be OOD. Otherwise it is more similar to classes in $\mathcal{C}_{\texttt{in}}$ and is inferred to be in-domain.

% In comparison to only using the first term as the score, the difference has better generalization to unseen OOD. Intuitively, for a test image from an unseen OOD class $\tilde{c} \notin \mathcal{C}_{\texttt{out}}$, the first term, the similarity to $\mathcal{C}_{\texttt{out}}$, can be small, but compared with the second term of the similarity to $\mathcal{C}_{\texttt{in}}$, it is still relatively big. Then $S_{\texttt{max\_logit\_diff}}(\vx)>0$, suggesting the image is OOD which is correct. 

% Thus our new score has better generalization to unseen OOD.
% We will discuss more in Section \ref{sec:why} why taking the difference between the similarity to $\mathcal{C}_{\texttt{out}}$ and $\mathcal{C}_{\texttt{in}}$, i.e. using the second term as a reference, is a better score than using the first term only.

Though $S_{\texttt{-max\_in\_prob}}(\vx)$ does not explicitly use the difference between the probability in $\mathcal{C}_{\texttt{in}}$ and that in $\mathcal{C}_{\texttt{out}}$, the $\mathrm{softmax}$ normalization actually considers the difference between the probability of \texttt{in} and that of the rest classes, including the OOD class with the maximum probability. Therefore, the two scores $S_{\texttt{-max\_in\_prob}}(\vx)$ and $S_{\texttt{max\_logit\_diff}}(\vx)$ measure similar quantities. 
In Section \ref{sec:why} we show the connection between the two. 

In summary, our proposed scores along with the baseline methods are listed in Table \ref{tab:eq_list}. Note that the proposed scores are computed only based on $\mathcal{C}_{\texttt{in}}$ and $\mathcal{C}_{\texttt{out}}$, not $\mathcal{C}_{\texttt{in}}^\prime$ and $\mathcal{C}_{\texttt{out}}^\prime$. $\mathcal{C}_{\texttt{in}}^\prime$ and $\mathcal{C}_{\texttt{out}}^\prime$ are only used at the test time for evaluating the performance on unknown. 

\begin{table}[h]
\centering
\caption{Comparison between the proposed scores and the baseline methods. 
% $p(c|\vx, \mathcal{C}_{\texttt{in}}) = \frac{e^{w_c}}{\sum_{j\in \mathcal{C}_{\texttt{in}}} e^{w_j}}$,
% $p(c| \vx, \mathcal{C}_{\texttt{in}} \cup \mathcal{C}_{\texttt{out}}) = \frac{e^{w_c}}{\sum_{j\in\mathcal{C_{\texttt{in}}}} e^{w_j} + \sum_{k\in\mathcal{C_{\texttt{out}}}} e^{w_k}}$,
% $w_c =  \vz_\texttt{img} \cdot \vz_\texttt{txt}^c$
}
\label{tab:eq_list}
\begin{tabular}{lccc}
\toprule
              & Uses $\mathcal{C}_{\texttt{out}}$ & Uses $\mathrm{softmax}$ & Definition \\ \hline
$S_{\texttt{-max\_prob}}$ \cite{ming2022delving}    & No          & Yes         &     $- \max_{c\in\mathcal{C}_{\texttt{in}}} p(c | \vx, \mathcal{C}_{\texttt{in}})$      \\
$S_{\texttt{sum\_out\_prob}}$  \cite{fort2021exploring,esmaeilpour2022zero}    & Yes         & Yes         &   $\sum_{c\in\mathcal{C_{\texttt{out}}}} p(c| \vx, \mathcal{C}_{\texttt{in}} \cup \mathcal{C}_{\texttt{out}})$       \\
$S_{\texttt{max\_out\_prob}}$   & Yes         & Yes         &   $\max_{c\in\mathcal{C_{\texttt{out}}}} p(c| \vx, \mathcal{C}_{\texttt{in}} \cup \mathcal{C}_{\texttt{out}})$       \\
$S_{\texttt{-max\_in\_prob}}$ (ours)  & Yes         & Yes         &    $- \max_{c\in\mathcal{C_{\texttt{in}}}} p(c| \vx, \mathcal{C}_{\texttt{in}} \cup \mathcal{C}_{\texttt{out}})$      \\
$S_{\texttt{max\_logit\_diff}}$ (ours) & Yes         & No          &   $\max_{d\in\mathcal{C_{\texttt{out}}}} w_d - \max_{c\in\mathcal{C_{\texttt{in}}}} w_c$      \\ \bottomrule \\
\multicolumn{4}{l} {{where $p(c|\vx, \mathcal{C}_{\texttt{in}}) = \frac{e^{w_c}}{\sum_{j\in \mathcal{C}_{\texttt{in}}} e^{w_j}}$,
$p(c| \vx, \mathcal{C}_{\texttt{in}} \cup \mathcal{C}_{\texttt{out}}) = \frac{e^{w_c}}{\sum_{j\in\mathcal{C_{\texttt{in}}}} e^{w_j} + \sum_{k\in\mathcal{C_{\texttt{out}}}} e^{w_k}}$,
$w_c =  \vz_\texttt{img} \cdot \vz_\texttt{txt}^c$.}}
\end{tabular}
\end{table}

\subsection{Extension to customized in- and out-of-distribution label sets}
To compute our score, one only needs a definition of $\mathcal{C}_{\texttt{out}}$ and $\mathcal{C}_{\texttt{in}}$. 
In fact we can extend $\mathcal{C}_{\texttt{out}}$ and $\mathcal{C}_{\texttt{in}}$ to be any label sets that are mutually exclusive. 
For example, the label sets can be defined at different hierarchical levels.
One can use the high level super class names to define $\mathcal{C}_{\texttt{out}}$ and $\mathcal{C}_{\texttt{in}}$. For example, for one-class person OOD detection, $\mathcal{C}_{\texttt{in}}$=\{person\}, $\mathcal{C}_{\texttt{out}}$=\{animals, cars, \dots\}. 
If the fine-grained level classes are known, one can define $\mathcal{C}_{\texttt{out}}$ and $\mathcal{C}_{\texttt{in}}$ in a more precise way. For example, $\mathcal{C}_{\texttt{in}}$=\{children, adults, \dots\}, $\mathcal{C}_{\texttt{out}}$=\{dogs, cats, trucks, buses, \dots\}. 
Because our text-image models can take any natural language as the text input, one can even use natural language to describe the sets via customized prompts. For example, `\emph{A photo of a \{class\} \{doing\}}'.
% $\mathcal{C}_{\texttt{out}}$=\{this is a photo of an animal, \dots\}.
Therefore, our method can be easily extended to any customized definitions of in- and out-of-distribution.

\subsection{One-class OOD detection in mixed in- and out-of-distribution multi-object images}
Detecting OOD in a mixed image that contains both in-domain and OOD objects is challenging. 
The OOD score for the image can be low due to the confounding of in-domain objects, causing false negatives in OOD detection. For example, an image of a person with a pet, or a dog on a chair.
To better detect the in-domain and OOD mixed images, we need to first detect the multiple objects in the images.
% and then remove the effect of the in-domain objects. 
Grounded-DINO \cite{liu2023grounding, kirillov2023segment} is one of the most powerful open-vocabulary detection model for detecting objects that correspond to an input text. 
For an image and a list of text labels, the output of Grounded-DINO contains a list of bounding boxes and the confidence scores for each box for each text label, i.e. $\texttt{score}_{(i, j)}$ for $i$-th bounding box and $j$-th text label. 

ImageNet-1K is known to contain multi-object images \cite{shankar2019evaluating,vasudevan2022does}. We apply Grounded-DINO to the images in ImageNet-Multilabel dataset.
We use $\mathcal{C}_{\texttt{out}}$ and $\mathcal{C}_{\texttt{in}}$ as the text input to Grounded-DINO.
The output $\texttt{score}_{(i, j)}$ for each bounding box are then treated as the logits, and we compute the OOD score for each bounding box.
% For an image containing multiple objects, we redefine the OOD score of the whole image as the maximum OOD scores among all the bounding boxes. 
To identify the in- and out- mixed images, we propose an mixture score $g(\vx)$, for indicating the confidence that the image contains both in-domain and OOD objects, as the greatest score difference among the bounding boxes.
\begin{align}
\label{eq:mixture}
    g(\vx) = \max_{b \in \texttt{boxes}} S_b - \min_{b \in \texttt{boxes}} S_b
\end{align}
where $S$ can be any of the scores in Table \ref{tab:eq_list}. An in- and out- mixed image will have $g(\vx)$ high.

\subsection{The connection between $S_{\texttt{-max\_in\_prob}}$ and $S_{\texttt{max\_logit\_diff}}$}
\label{sec:why}
In this section we show that the score $S_{\texttt{-max\_in\_prob}}$ and $S_{\texttt{max\_logit\_diff}}$ share similar components. 
To unify the two, we first take the logarithm of the softmax score. Since logarithm is a monotonic function, the transformation preserves the order of the values. 
Since $S_{\texttt{-max\_in\_prob}}=-\max_{c\in\mathcal{C_{\texttt{in}}}} p(c| \vx, \mathcal{C}_{\texttt{in}} \cup \mathcal{C}_{\texttt{out}})<0$, to take the logarithm we reverse its value, 
$\log \max_{c\in\mathcal{C_{\texttt{in}}}} p(c| \vx, \mathcal{C}_{\texttt{in}} \cup \mathcal{C}_{\texttt{out}}) = \max_{c\in\mathcal{C_{\texttt{in}}}} w_c - \log \left( \sum_{j\in\mathcal{C_{\texttt{in}}}} e^{w_j} + \sum_{k\in\mathcal{C_{\texttt{out}}}} e^{w_k} \right)$. 
The second term can be decomposed into the the sum of $\max_{q\in\mathcal{C_{\texttt{out}}}}e^{w_q}$ and the rest, thus $\log \left( \sum_{j\in\mathcal{C_{\texttt{in}}}} e^{w_j} + \sum_{k\in\mathcal{C_{\texttt{out}}}} e^{w_k} \right) = \log \left( \max_{q\in\mathcal{C_{\texttt{out}}}}e^{w_q} \left( 1 + r \right) \right) = \max_{q\in\mathcal{C_{\texttt{out}}}}w_q + \log \left( 1 + r \right) $, where $r = \frac{\sum_{l\in\mathcal{C_{\texttt{in}}} \cup \mathcal{C_{\texttt{out}}}, l\neq q} e^{w_l}}{\max_{q\in\mathcal{C_{\texttt{out}}}}e^{w_q}} $. Then we have 
% \begin{small}
% \begin{align*}
    $- \log \max_{c\in\mathcal{C_{\texttt{in}}}} p(c| \vx, \mathcal{C}_{\texttt{in}} \cup \mathcal{C}_{\texttt{out}}) =  \max_{d\in\mathcal{C_{\texttt{out}}}}w_d - \max_{c\in\mathcal{C_{\texttt{in}}}}w_c + \log \left( 1 + r \right) = S_{\texttt{max\_logit\_diff}}(\vx) + \log \left( 1 + r \right).$
% \end{align*}
% \end{small}
% As we can see, the first component is the same as $S_{\texttt{max(out) - max(in)}}(\vx)$. 
% Thus $S_{\texttt{-max\_in\_prob}} = - \exp\left(-\left(S_{\texttt{max\_logit\_diff}}+\log(1+r)\right)\right)$
The ratio $r$ measures the peakiness of predicted probability distribution. When the predicted probability distribution is concentrated at the predicted OOD class, $r\approx 0$ and thus $ \log(-S_{\texttt{-max\_in\_prob}}) \approx -S_{\texttt{max\_logit\_diff}}$.
When the predicted probability distribution is spread out over a wide range of values, $r$ is large and then $\log(-S_{\texttt{-max\_in\_prob}}) < -S_{\texttt{max\_logit\_diff}}$.
A greater OOD score favors OOD but not in-domain samples. So depending on the use case, the two scores can have different advantages. 

\section{Experimental evaluation}

We evaluate our proposed methods, along with a few baseline methods, on large-scale datasets and real world challenging problems. 
% Unlike the previous studies \cite{fort2021exploring,ming2022delving} which either use the small-scale datasets like CIFAR or only consider images without distributional shift.
% in this work, 
% We are interested in evaluating the tasks that are more challenging but very common in real-world settings. 
Here are some of the challenges we considered:
\vspace{-0.5em}
\begin{itemize}[leftmargin=1.5em,itemsep=-0.1em]
    \item Unseen classes: We evaluate the scenarios where the test images belong to none of the classes in $\mathcal{C}_{\texttt{out}}$ and $\mathcal{C}_{\texttt{in}}$.
    For example, for the problem of person detection, we set $\mathcal{C}_{\texttt{out}}=$\{animal, car, food, $\dots$\}, but at the test time we have images of $\mathcal{C}_{\texttt{out}}^\prime=$\{toy, tree, \dots\}, $\mathcal{C}_{\texttt{out}} \cap \mathcal{C}_{\texttt{out}}^\prime = \emptyset$. Similarly, we can have $\mathcal{C}_{\texttt{in}}^\prime$=\{infant, senior, \dots\}, which are not seen in $\mathcal{C}_{\texttt{in}}$=\{children, adults, \dots\}.
    \item Distributional shift: We evaluate the scenarios where there is a covariate shift in inputs while the conditional distribution of classes is unchanged \cite{sugiyama2012machine}. 
    For example, a drawing of a person is a shift from the natural person images. The distributional shift datasets we evaluate with are ImageNet-V2 \cite{recht2019imagenet}, ImageNet-A \cite{hendrycks2019nae}, ImageNet-R \cite{hendrycks2020many}, and ImageNet-Sketch \cite{wang2019learning}.
    % ImageNet-Sketch\footnote{\url{https://www.tensorflow.org/datasets/catalog/imagenet_sketch}}, ImageNet-A\footnote{\url{https://www.tensorflow.org/datasets/catalog/imagenet_a}}, ImageNet-R\footnote{\url{https://www.tensorflow.org/datasets/catalog/imagenet_r}}, and ImageNet-V2\footnote{\url{https://www.tensorflow.org/datasets/catalog/imagenet_v2}}.
    \item Multi-object images: We evaluate on images that contain a mixture of in-domain and OOD objects, using ImageNet-Multilabel dataset \cite{shankar2019evaluating,vasudevan2022does}.
    % \footnote{\url{https://www.tensorflow.org/datasets/catalog/imagenet2012_multilabel}}. 
    In the real world, images may contain multiple objects, sometimes a mixture of in-domain and OOD objects. For example, a person with a dog (non-person). Those in- and out- mixed images are hard examples for OOD detection. 
\end{itemize}
\vspace{-0.5em}
\paragraph{Datasets}

% Originally we wanted to study the problem of detecting \texttt{non-person} images in the \texttt{person} recognition system. However, because human images can be sensitive, we decided to use the problems of detecting 
% (1) \texttt{non-dogs versus dogs}, and (2) \texttt{non-birds versus birds}, as proxies to the original problem. 
We evaluate our model on the ImageNet-1K dataset validation split \cite{ILSVRC15}. We group the images in the dataset by their class labels, following the Pascal Visual Object Classes (VOC) WordNet hierarchy \cite{miller1995wordnet,li2022bigdatasetgan}.
The Pascal VOC provides a mapping from the ImageNet-1K classes to a few common superclasses such as dog, cat, bird, etc. 
The number of subclasses in each superclass is as follows, \{dog: 118, bird: 59, boat: 6, bottle: 7, bus: 3, car: 10, cat: 7, chair: 4, diningtable: 1, horse: 1, person: 3, sheep:1, train: 1, aeroplane: 1, bicycle: 2\}, in total 224 classes. The remaining 776 classes are from rare categories such as ``fox squirrel'', ``snow leopard'', ``cowboy hat'', ``electric guitar'', forming a long-tail distribution. None of the 776 classes are in the common categories. 

% To mimic the scenario of non-person anomaly detection in person identification system, in the task of non-dog vs dog, we want to detect the anomalies, i.e. the non-dog images, before sending the images to a dog classifier.
Based on the class hierarchy, we evaluate the one-class OOD detection problems for the superclasses dog, bird, bus, car, cat, chair, person individually. 
For each of the one-class OOD problem, we use the images of classes belonging to the superclass as the in-domain data, and the images of the rest classes as the OOD data. 
Since we would like to evaluate the unseen classes, the set of in-domain classes are randomly split into equal size non-overlapping $\mathcal{C}_{\texttt{in}}$ and $\mathcal{C}_{\texttt{in}}^\prime$. 
To make the results reproducible, we here use the first half of classes as  $\mathcal{C}_{\texttt{in}}$ and the second half as $\mathcal{C}_{\texttt{in}}^\prime$.
We use the OOD classes belonging to the common super categories as $\mathcal{C}_{\texttt{out}}$, and the OOD classes belonging to the remaining 776 classes as $\mathcal{C}_{\texttt{out}}^\prime$.
For example, for the dog vs non-dog problem, we split the 118 dog classes into $\mathcal{C}_{\texttt{in}}$ and $\mathcal{C}_{\texttt{in}}^\prime$. $\mathcal{C}_{\texttt{out}}$ consists of the classes belonging to the common categories \{bird, boat, bottle, bus, \dots, person, chair\}, and $\mathcal{C}_{\texttt{out}}^\prime$  consists of the 776 rare classes.

To compute our scores, we assume we only have $\mathcal{C}_{\texttt{in}}$ and $\mathcal{C}_{\texttt{out}}$.
To evaluate the performance on OOD detection, we consider how well the scores can separate in-domain and OOD images belonging to (1) $\mathcal{C}_{\texttt{in}}$ vs $\mathcal{C}_{\texttt{out}}$,
% (2) $\mathcal{C}_{\texttt{in}}$ vs $\mathcal{C}_{\texttt{out}}^\prime$, (3) $\mathcal{C}_{\texttt{in}}^\prime$ vs $\mathcal{C}_{\texttt{out}}$, 
and (2) $\mathcal{C}_{\texttt{in}}^\prime$ vs $\mathcal{C}_{\texttt{out}}^\prime$. 
% (2) evaluates the OOD performance on images from unseen classes. 
To evaluate the performance on distribution shifted data, we use ImageNet-V2, ImageNet-A, ImageNet-R, and ImageNet-Sketch in the corresponding classes to construct the test data. 

Besides the OOD detection using the superclasses as in-domain, we also consider a narrower in-domain OOD detection problem. The goal is to show that our method is general for both wide and narrow one-class problem.
We use ``terrier'', a dog sub-type, to construct the this narrower one-class OOD detection problem. 
Among the 118 dog classes, 23 of them are terrier. 
We randomly split the 23 classes into non-overlapping $\mathcal{C}_{\texttt{in}}$ and $\mathcal{C}_{\texttt{in}}^\prime$.
Again $\mathcal{C}_{\texttt{out}}$ consists of the classes belonging to the common categories, and $\mathcal{C}_{\texttt{in}}^\prime$ consists of the 776 rare classes. 
The rest 95 non-terrier dog classes are considered as near-OOD $\mathcal{C}_{\texttt{in}}^{\texttt{near}}$.

\textbf{Evaluation metrics} 
To evaluate the performance on one-class OOD detection, we use the area under the ROC curve (AUROC) between the scores of in-domain and that for OOD. The higher the AUROC score suggests a better separation between in- and out-of-distribution. 

\textbf{Models} 
We use CLIP ViT-B/16 as the main model for evaluating our method. We did ablation study on ViT-L/14, and the conclusions were consistent (see Table S\ref{tab:inet-vitl}).
\vspace{-0.5em}

% \subsection{Our score outperforms the baseline on detecting anomaly images from unseen classes}
\subsection{Our scores outperform the baselines on one-class OOD detection tasks} \label{sec:results_compare_methods}
We evaluate the performance of the listed scores in Table \ref{tab:eq_list} on one-class OOD detection tasks using ImageNet and its variants. We are in particular interested in the generalization ability of the scores on unseen samples, and distribution shifted samples.
As shown in Table \ref{tab:joint_table}, evaluated on the tasks of dog vs non-dog, car vs non-car, and person vs non-person, our proposed scores $S_{\texttt{-max\_out\_prob}}$ and $S_{\texttt{max\_logit\_diff}}$ consistently outperform the baselines, achieving the highest AUROCs. The two proposed scores have similar high performance. For all the three tasks, our scores' AUC approaches 1.0 on the images from pre-defined classes $\mathcal{C}_{\texttt{in}}$ and $\mathcal{C}_{\texttt{out}}$. 
It is more challenging to detect images from the unseen classes $\mathcal{C}_{\texttt{in}}^\prime$ and $\mathcal{C}_{\texttt{out}}^\prime$. 
The most challenging task is person vs non-person detection on unseen data, with the best AUC only 0.78 on ImageNet. 
That is possibly because (1) the person superclass consists of a set very diverse subtypes so that having one subtype (such as a groom) in $\mathcal{C}_\texttt{in}$ does not help much on detecting another subtype (such as a scuba diver). (2) There are only 3 person classes \{baseball player, groom, scuba diver\} in ImageNet so that $\mathcal{C}_{\texttt{in}}$ has very limited coverage of the in-domain. Including more person class names in $\mathcal{C}_{\texttt{in}}$ would help to reduce the ambiguity between in-domain and OOD. 

% has only two classes which has limited help on the decision (where our method is not the best).

Our scores also perform the best on other ImageNet variant datasets, ImageNet-V2, ImageNet-R, ImageNet-A, and ImageNet-Sketch. 
ImageNet-A in appears to be the most difficult dataset, since even our best score has 0.06 and 0.09 drop on AUC compared with ImageNet. 
The proposed two scores perform similarly well on most of the tasks, except for ImageNet-A. 
$S_{\texttt{-max\_in\_prob}}$ has AUC 0.1 higher than $S_{\texttt{max\_logit\_diff}}$ on detecting unseen samples for the car vs non-car task. 
On the other hand, $S_{\texttt{max\_logit\_diff}}$ is better than $S_{\texttt{-max\_in\_prob}}$ by 0.08 AUC on detecting dog vs non-dog. 
% One notable difference is in the person vs non-person task, where $S_{\texttt{max\_logit\_diff}}$ (ours) has a much higher AUC (0.9818) than $S_{\texttt{-max\_in\_prob}}$ (0.6337) on detecting unseen in-domain $\mathcal{C}_{\texttt{in}}^\prime$ vs $\mathcal{C}_{\texttt{out}}$.

Note that $S_{\texttt{-max\_out\_prob}}$ does not perform as well as $S_{\texttt{-max\_in\_prob}}$, especially on the unseen OOD. That is because $\mathcal{C}_{\texttt{out}}$ does not cover all the possible categories in OOD space. An unseen OOD that is not similar to any $\mathcal{C}_{\texttt{out}}$ will have the $S_{\texttt{-max\_out\_prob}}$ small.
We also evaluate the performance on other one-class tasks such as bird vs non-bird. Please find the full table in Tables~S\ref{tab:inet}-\ref{tab:inet-s} .

\begin{table}[htb]
%\vspace{-1em}
\caption{One-class OOD detection across datasets for various in-domain cases evaluated using AUC$\uparrow$. Our scores consistently outperform the baselines for detecting samples from unseen classes and under distribution shift. Note that ImageNet-A does not have person images (N/A in Table below).}
\label{tab:joint_table}
\centering
\resizebox{\textwidth}{!}{
\begin{tabular}{l|l|cc|cc|cc}
\toprule

\multicolumn{2}{c|}{\multirow{2}{*}{ }} & \multicolumn{2}{c|}{Dog vs non-dog} & \multicolumn{2}{c|}{Car vs non-car} & \multicolumn{2}{c}{Person vs non-person} \\
\cline{3-8}
\multicolumn{1}{l}{} &  & $\mathcal{C}_{\texttt{in}}$ vs $\mathcal{C}_{\texttt{out}}$ &  $\mathcal{C}_{\texttt{in}}^\prime$ vs $\mathcal{C}_{\texttt{out}}^\prime$  & $\mathcal{C}_{\texttt{in}}$ vs $\mathcal{C}_{\texttt{out}}$ &  $\mathcal{C}_{\texttt{in}}^\prime$ vs $\mathcal{C}_{\texttt{out}}^\prime$  & $\mathcal{C}_{\texttt{in}}$ vs $\mathcal{C}_{\texttt{out}}$ &  $\mathcal{C}_{\texttt{in}}^\prime$ vs $\mathcal{C}_{\texttt{out}}^\prime$ \\
%\hline
\midrule
\multirow{4}{*}{\begin{turn}{270} \begin{tabular}{@{}c@{}}ImageNet \\ \color{white} * \end{tabular} \end{turn}} 
& $S_{\texttt{-max\_prob}}$ &  0.9687   &  0.8357   &  0.8209  & 0.5209 & 0.7781 & 0.5096  \\
& $S_{\texttt{sum\_out\_prob}}$ &  0.7715    &    0.8128    &  0.8957 & 0.7440 &  0.9859 &0.6605 \\
& $S_{\texttt{max\_out\_prob}}$ &  0.9971   &  0.7321   & 0.9723  & 0.3652 & 0.9885& \textbf{0.7803}  \\
& $S_{\texttt{-max\_in\_prob}}$ (ours) &   \underline{0.9979}  &    \underline{0.9847} & \underline{0.9944}   &  \textbf{0.9835}&  \underline{0.9995} &  0.4900 \\ 
& $S_{\texttt{max\_logit\_diff}}$ (ours) &    \textbf{1.0000}  &   \textbf{0.9896} &   \textbf{0.9996}   & \underline{0.9360} & \textbf{0.9997}&     \underline{0.6974}\\  
            \hline
% & $S_{\texttt{-max\_prob}}$ &  0.7962   &   0.8647  & 0.8886   & 0.6446 & 0.9968 &   0.3288\\
% & $S_{\texttt{sum\_out\_prob}}$ &  0.9716    &   0.8250     &0.8065   & 0.5237 &0.7684   & 0.5696\\
% & $S_{\texttt{max\_out\_prob}}$ &   0.9972    &    0.7240 &0.9702   & 0.4379 &0.9920 &   0.5632\\
% & $S_{\texttt{-max\_in\_prob}}$ (ours) &   \underline{0.9983}  &    \underline{0.9898} & \underline{0.9951}   &  \textbf{0.9796}&  \textbf{0.9999} &  \underline{0.6376} \\ 
% & $S_{\texttt{max\_logit\_diff}}$ (ours) &    \textbf{1.0000}  &   \textbf{0.9933} &   \textbf{0.9997}   & \underline{0.9737} & \underline{0.9998}&     \textbf{0.6393}\\  
%             \hline
\multirow{4}{*}{\begin{turn}{270}\begin{tabular}{@{}c@{}}ImageNet \\ v2\end{tabular}\end{turn}} 
& $S_{\texttt{-max\_prob}}$& 0.7163  & 0.7384 & 0.9081 &  0.6649  & 0.9589 &  0.7131  \\
& $S_{\texttt{sum\_out\_prob}}$ & 0.9590 &  0.8083  & 0.8570 & 0.4885 & 0.7567& 0.4104  \\
& $S_{\texttt{max\_out\_prob}}$ & 0.9923 & 0.7067 & 0.9449 &  0.3651   & 0.9726 &   \underline{0.6993}\\
& $S_{\texttt{-max\_in\_prob}}$ (ours) & \underline{0.9945}  & \underline{0.9795}& \underline{0.9870} &   \textbf{0.9729}& \underline{0.9975} & 0.6436 \\ 
& $S_{\texttt{max\_logit\_diff}}$ (ours) &   \textbf{0.9994} &    \textbf{0.9836} &   \textbf{0.9988} &  \underline{0.9272} &   \textbf{0.9997} &   \textbf{0.7172}  \\ 
\hline

\multirow{4}{*}{\begin{turn}{270} \begin{tabular}{@{}c@{}}ImageNet\\ R  \color{white} * \end{tabular} \end{turn}}
& $S_{\texttt{-max\_prob}}$&  0.8148   & 0.6334    &  0.8902   &  0.5690   & 0.9723   &   0.4824  \\
&  $S_{\texttt{sum\_out\_prob}}$ &  0.9616 &  0.7812      &  0.9247  &   0.3688   &   0.8639   &  0.5818  \\
& $S_{\texttt{max\_out\_prob}}$ & 0.9758 &  0.6950   &  0.9584  &   0.2464  &  0.9662  &  0.5917\\
& $S_{\texttt{-max\_in\_prob}}$ (ours) &  \underline{0.9903}&  \underline{0.9733}   & \underline{0.9976}   &      \textbf{0.9420}   &   \underline{0.9979}    & \underline{0.5924} \\  
& $S_{\texttt{max\_logit\_diff}}$ (ours) &    \textbf{0.9990} &     \textbf{0.9726}  &   \textbf{0.9998}    &   \underline{0.8177}   &    \textbf{0.9990}    &     \textbf{0.6300} \\ 
\hline

\multirow{4}{*}{\begin{turn}{270}\begin{tabular}{@{}c@{}}ImageNet \\ Adversarial\end{tabular}\end{turn}}  
& $S_{\texttt{-max\_prob}}$&  0.3544&  0.3991   &  0.8668   &   0.6388  & N/A &  N/A \\
& $S_{\texttt{sum\_out\_prob}}$ &  \underline{0.9399}&    0.6848   &  0.8528     &   0.3853   & N/A &  N/A \\
& $S_{\texttt{max\_out\_prob}}$ &  0.9307 &  0.6778   &  0.8486  &    0.3886   & N/A & N/A \\
& $S_{\texttt{-max\_in\_prob}}$ (ours) &  0.8963 &   \underline{0.9261}  & \underline{0.9792}    &    \textbf{0.8860}   & N/A & N/A\\ 
& $S_{\texttt{max\_logit\_diff}}$ (ours) &    \textbf{0.9769}   &   \textbf{0.9333}   &    \textbf{0.9935}  &  \underline{0.7880}  & N/A & N/A \\

\hline
\multirow{4}{*}{\begin{turn}{270} \begin{tabular}{@{}c@{}}ImageNet \\ Sketch\end{tabular} \end{turn}} 
& $S_{\texttt{-max\_prob}}$ &  0.7676   &0.8073  &  0.9179   &   0.6674 &  0.9678   &  0.3821  \\
& $S_{\texttt{sum\_out\_prob}}$ & 0.9643  & 0.8232 &  0.9150   &  0.4945  &   0.7934   &   0.5635   \\
& $S_{\texttt{max\_out\_prob}}$ &  0.9820  & 0.6979   & 0.9487   &  0.3672  &  0.9869  & 0.6203\\
& $S_{\texttt{-max\_in\_prob}}$ (ours) & \underline{0.9851}   &    \textbf{0.9868}   &  \underline{0.9967}  &   \textbf{0.9785}    & \underline{0.9985}  & \underline{0.6528}\\  
& $S_{\texttt{max\_logit\_diff}}$ (ours) &    \textbf{0.9993}   &  \underline{0.9850}  &    \textbf{0.9993}   &  \underline{0.9199}  &   \textbf{0.9999}   &   \textbf{0.6790} \\  
\bottomrule

\end{tabular}}
\end{table}

\subsection{Customized in-domain and OOD label sets help to improve performance}

To compute our scores, one only needs the two set of labels $\mathcal{C}_{\texttt{in}}$ and  $\mathcal{C}_{\texttt{out}}$. 
The label set can be at different hierarchical levels, or even include natural languages. 
Here we explore a narrow in-domain OOD detection problem, terrier (a sub-type of dog) vs non-terrier, to demonstrate that customized label sets help to improve the performance. 
This problem also helps us to evaluate the performance on near-OOD, since the non-terrier dogs are naturally defined near-OOD.
Both $\mathcal{C}_{\texttt{in}}$ and $\mathcal{C}_{\texttt{out}}$ can be defined at coarse- or fine-grained levels.
The coarse level $\mathcal{C}_{\texttt{in}}^{\texttt{coarse}}$=\{terrier\} or the fine-grained level $\mathcal{C}_{\texttt{in}}^{\texttt{fine}}$=\{Boston terrier, Norwich terrier, \dots \} is paired with 
coarse level $\mathcal{C}_{\texttt{out}}^{\texttt{coarse}}$=\{bird, boat, \dots'\} or fine-grained level $\mathcal{C}_{\texttt{out}}^{\texttt{fine}}$=\{robin, \dots, canoe, \dots \} to compute our scores.
To better detect near-OOD, we also consider adding near-OOD classes to $\mathcal{C}_{\texttt{out}}^{\texttt{near}}$.
As shown in Table \ref{tab:terrier}, providing more precise fine-grained level labels helps to improve the performance for both scores. 
Adding the near-OOD class labels further helps to improve the performance, particularly on near-OOD.

\begin{table}[ht]
\centering
\caption{One-class OOD detection for dog sub-type terrier using different $\mathcal{C}_{\texttt{in}}$ and $\mathcal{C}_{\texttt{out}}$ label sets. More fine-grained label sets help to improve the performance for both scores.
% The score is based on Eq. (\ref{eq:max_ind_softmax_all}).
}
\label{tab:terrier}
\resizebox{0.8\textwidth}{!}{
\begin{tabular}{llllll}
%  &  \multicolumn{5}{c}{Test set splits} \\ \toprule
Label sets                     & $\mathcal{C}_{\texttt{in}}$ vs $\mathcal{C}_{\texttt{out}}$ & $\mathcal{C}_{\texttt{in}}^\prime$ vs $\mathcal{C}_{\texttt{out}}^\prime$ & $\mathcal{C}_{\texttt{in}}$ vs $\mathcal{C}_{\texttt{out}}^{\texttt{near}}$ & $\mathcal{C}_{\texttt{in}}^\prime$ vs $\mathcal{C}_{\texttt{out}}^{\texttt{near}}$ & Average \\ \toprule
\multicolumn{6}{c}{$S_{\texttt{-max\_in\_prob}}$ }    \\ \midrule
$\mathcal{C}_{\texttt{in}}^{\texttt{coarse}} \cup \mathcal{C}_{\texttt{out}}^{\texttt{coarse}}$ & 0.9967                 & 0.9982                    & 0.8333               & 0.8722                 & 0.9251   \\
$\mathcal{C}_{\texttt{in}}^{\texttt{fine}} \cup \mathcal{C}_{\texttt{out}}^{\texttt{coarse}}$        & 0.9954                 & 0.9960                    & 0.8770               & 0.7939                 & 0.9156   \\
$\mathcal{C}_{\texttt{in}}^{\texttt{fine}} \cup \mathcal{C}_{\texttt{out}}^{\texttt{fine}}$      & 0.9998                 & 0.9992                    & 0.9051               & 0.8584                 & 0.9406   \\
\ +$\mathcal{C}_{\texttt{out}}^{\texttt{near}}$          & 0.9984                 & 0.9974                    & 0.9205               & 0.8704                 & \underline{0.9467}   \\
 \ \ \ + add 80 prompts       & 0.9986                 & 0.9981                    & 0.9243               & 0.8687                 & \textbf{0.9474}  \\
\ \ \ + add actions          & 0.9981                 & 0.9977                    & 0.9196               & 0.8710                 & 0.9466   \\ \midrule
\multicolumn{6}{c}{$S_{\texttt{max\_logit\_diff}}$ }    \\ \midrule
$\mathcal{C}_{\texttt{in}}^{\texttt{coarse}} \cup \mathcal{C}_{\texttt{out}}^{\texttt{coarse}}$ & 0.9994 & 0.9956 & 0.8170 & 0.8390 & 0.9128 \\
$\mathcal{C}_{\texttt{in}}^{\texttt{fine}} \cup \mathcal{C}_{\texttt{out}}^{\texttt{coarse}}$       & 1.0000 & 0.9987 & 0.9036 & 0.8613 & 0.9409 \\
$\mathcal{C}_{\texttt{in}}^{\texttt{fine}} \cup \mathcal{C}_{\texttt{out}}^{\texttt{fine}}$       & 1.0000 & 0.9995 & 0.9072 & 0.8767 & 0.9458 \\
\ +$\mathcal{C}_{\texttt{out}}^{\texttt{near}}$        & 0.9997 & 0.9480 & 0.9758 & 0.9327 & \textbf{0.9640} \\
\ \ \ + add 80 prompts      & 0.9998 & 0.9411 & 0.9769 & 0.9346 & 0.9631 \\
\ \ \ + add actions         & 0.9998 & 0.9400 & 0.9779 & 0.9350 & \underline{0.9632} \\ \bottomrule
\end{tabular}
}
\end{table}

Since the CLIP model can input any form of text, natural language that describes the in-domain and OOD classes can also be used for computing our scores. 
A simple way to generate sentences from the class names is to use the prompt template, such as `\emph{A photo of \{\}}'.
We apply the 80 hand-crafted prompts of \cite{radford2021learning} to each of the class name in the label set, and take the average of their embeddings per class name as the representation of that class. 
We also consider adding dogs' actions such as playing, running, sleeping, and walking, using the prompt template `\emph{A photo of a \{class\} \{doing\}}'.
The results show that adding the 80 prompts could improve the performance further for $S_{\texttt{-max\_in\_prob}}$. However, adding actions does not have much effect on AUC.

\subsection{OOD detection in mixed in-domain and OOD multi-object images}

Images with mixed in-domain and OOD objects are difficult to detect because the in-domain object can lower the OOD score, causing the images to be misclassified. 
We aim to flag those mixed images such that possible post-processing can be executed to correctly classify those images.

\begin{figure}[h]
% \vspace{-10pt}
\centering
\includegraphics[width=0.9\textwidth]{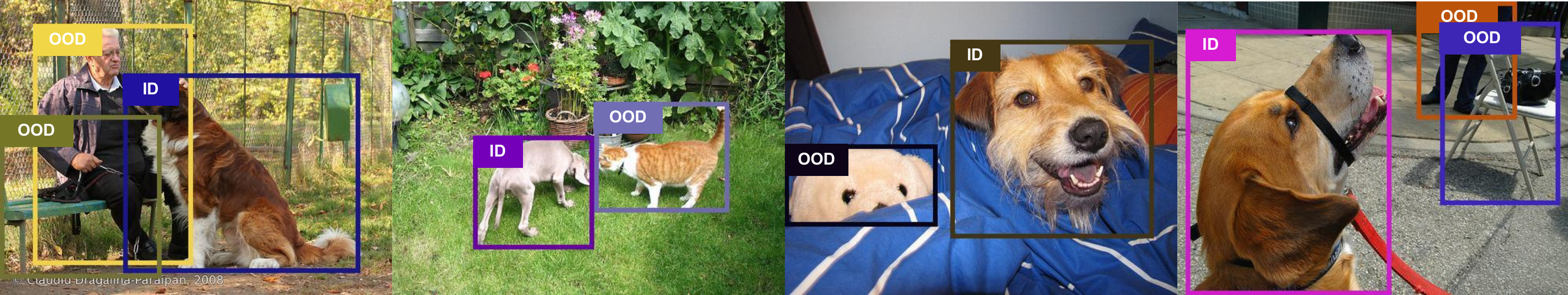}
\caption{Our methods detect OOD at bounding box level. Images having mixture of in-domain and OOD objects are identified.}
\label{fig:multilabel}
\end{figure}

We use the ImageNet-Multilabel dataset \cite{shankar2019evaluating,vasudevan2022does} to evaluate the performance of OOD detection in mixed in-domain and OOD images. 
There are 1743 ImageNet images with more than one bounding box prediction. We want to identify images that contain both in-domain and OOD objects  (Figure \ref{fig:multilabel}).

As shown in the bottom section of Table \ref{tab:multilabel}, none of the single image OOD scores are able to identify the mixed in- and out-domain images, with AUCs between pure and mixed around 0.5.
% Since a single score for the whole image can be confounded by the in-domain object, 

\begin{table}[h]
% \vspace{-10pt}
\centering
\caption{Identifying in-domain and OOD mixed multi-object images using mixture score $g(\vx)$ defined based on different OOD scores. None of the single scores can identify mixed images. New scores based on bounding box detection improve the performance, and our scores outperform the baselines. 
% $p(c|\vx, \mathcal{C}_{\texttt{in}}) = \frac{e^{w_c}}{\sum_{j\in \mathcal{C}_{\texttt{in}}} e^{w_j}}$,
% $p(c| \vx, \mathcal{C}_{\texttt{in}} \cup \mathcal{C}_{\texttt{out}}) = \frac{e^{w_c}}{\sum_{j\in\mathcal{C_{\texttt{in}}}} e^{w_j} + \sum_{k\in\mathcal{C_{\texttt{out}}}} e^{w_k}}$,
% $w_c =  \vz_\texttt{img} \cdot \vz_\texttt{txt}^c$
}
\label{tab:multilabel}
\resizebox{1\textwidth}{!}{
\begin{tabular}{llcccc}
& & \multicolumn{2}{c}{Dog} & \multicolumn{2}{c}{Bird}  \\ \toprule
             & & Pure in vs mix & Pure OOD vs mix & Pure in vs mix & Pure OOD vs mix   \\ \midrule
\multirow{3}{*}{Scores using bbox} & $g(S_{\texttt{sum\_out\_prob}})$       &  0.6598     &  0.6918  & \underline{0.8557}  & 0.7844  \\
& $g(S_{\texttt{-max\_in\_prob}})$ (ours)  &  \underline{0.6836}   &   \textbf{0.9570}  & 0.7211 & \textbf{0.9833}\\
& $g(S_{\texttt{max\_logit\_diff}})$ (ours) &  \textbf{0.6861}    &  \underline{0.8672} &  \textbf{0.8846} & \underline{0.8460} \\ 
\toprule
\multirow{3}{*}{Single score} & $S_{\texttt{-max\_prob}}$     &   0.4907   & 0.5127  &  0.4869 & 0.4787 \\ 
& $S_{\texttt{sum\_out\_prob}}$&   0.5492   &  0.4940 &  0.4908 & 0.4990 \\ 
& $S_{\texttt{max\_out\_prob}}$  &   0.5445   &  0.5152 & 0.4998  & 0.4872 \\ 
& $S_{\texttt{-max\_in\_prob}}$ &   0.5527   & 0.5089  & 0.5169  & 0.4946 \\ 
& $S_{\texttt{max\_logit\_diff}}$  &   0.5794  &  0.5191 &  0.4886 & 0.4639 \\ \bottomrule

%               & Pure bird vs Mix & Pure non-bird vs Mix  \\ \toprule
% $S_{\texttt{sum\_out\_prob}}$      & 0.8557  & 0.7844 \\
% $S_{\texttt{-max\_in\_prob}}$ (ours)  & \underline{0.7211} & \textbf{0.9833} \\
% $S_{\texttt{max\_logit\_diff}}$ (ours) &  \textbf{0.8846} & \underline{0.8460} \\ \bottomrule \\
\end{tabular}
}
\end{table}

Image segmentation and object detection are needed for identifying those mixed images.
We use Grounding-DINO to localize the multiple objects in bounding boxes along with their confidence scores. 
Since  Grounding-DINO has a limitation on input text length, we decided to use the high level class names for $\mathcal{C}_{\texttt{in}}=$\{dog\} and $\mathcal{C}_{\texttt{out}}=$\{bird, boat, person, \dots\}.
As described in Section 2.4, for each bounding box, we have a list of confidence scores corresponding to the list of class names. Then we compute our proposed OOD scores separately for each bounding box. 
To find the mixed in-domain and OOD multi-object images, we define the mixture score $g(\vx)$ in Eq. (\ref{eq:mixture}) for each image as the greatest score difference among the bounding boxes.

% We use the ImageNet-Multilabel dataset \cite{shankar2019evaluating,vasudevan2022does} to evaluate the performance of OOD detection in mixed in-domain and OOD images. 
% There are 1743 ImageNet images with more than one bounding box prediction. We want to identify images that contain both in-domain and OOD objects. 
Table~\ref{tab:multilabel} shows our proposed mixture score $g(\vx)$, is able to distinguish the mixed images from pure in-domain and pure OOD images, for both dog and bird datasets, having AUCs higher than the baseline.

\section{Related work}

% \subsection{Out-of-distribution detection using text-image models}

\paragraph{One-class anomaly detection}

Anomaly detection can be formulated as a one-class classification problem \cite{khan2010survey}, which aims to learn the distribution of the normal data only, and then predicts anomalies as data points that are out of the normal distribution.
SVM based one-class classification, also called support vector data description (SVDD), fits a hypersphere with the minimum volume that includes most of the normal data points \cite{noumir2012simple}. 
DeepSVDD leverages the ability of deep neural networks to first learn a good representation before mapping the data to a hypersphere \cite{ruff2018deep}. 
% Another unsupervised approach is to use reconstruction error in the autoencoder as an anomaly score \cite{chalapathy2017robust}.  
Later works propose hybrid models that use an autoencoder to learn the data representation, and then map the representation to a one-class classification model \cite{chalapathy2018anomaly,zhang2021anomaly,hojjati2021dasvdd,park2021wrong}.
The existing one-class anomaly detection methods have limitations that (1) they need to be trained so they are not zero-shot, (2) they cannot leverage the abnormal data into training, and (3) they are evaluated on simple datasets such as MNIST and CIFAR-10 and on pre-defined closed OOD sets.

% and using simple datasets . The  where one of the class is set as in-domain and the rest are set as OOD (such as number 1s are in-domain and the number 2-9s are OOD). However,

\paragraph{Multi-label OOD detection}
To the best of our knowledge, all the existing work on multi-label OOD detection is to detect the images that contain none of the in-domain objects \cite{wangenergy,wang2022multi,sundar2020out}.
For example, if the in-domain classes are \{dog, cat\}, the goal is to detect images that do not contain any instances of dogs or cats, such as an image of a chair.
In comparison, our work aims to detect the images that contain a mixture of in-domain and OOD objects, such as a dog on a chair. 
It is challenging to detect the images with mixed in-domain and OOD objects, because the in-domain objects can confound the OOD score.
We believe our work is the first to address this problem for in- and out- mixed \textit{multi-object} out-of-distribution detection.

\section{Conclusion and discussion}
We propose a novel one-class open-set OOD detector that leverages text-image pre-trained models in a zero-shot fashion and incorporates various descriptions of in-domain and OOD.  Unlike prior work, we focus on more challenging and realistic settings for OOD detection. We evaluate on images that are from the long-tail of unseen classes, distribution shifted images, and in-domain and OOD mixed multi-object images. Our method is flexible enough to detect any types of OOD, defined with fine- or coarse-grained labels. Our method shows superior performance over previous baselines on all benchmarks.
Nonetheless, our method has room for more improvement. We are interested in additional ways to incorporate natural language to define in-domain and OOD beyond prompts. An additional question is how to effectively use negation to define OOD. Better understanding of the advantages and trade-offs of the proposed two scores is also part of the future work.

%%%%%%%%%%%%%%%%%%%%%%%%%%%%%%%%%%%%%%%%%%%%%%%%%%%%%%%%%%%%

\section*{Acknowledgements}
We thank Neil Houlsby and Sharat Chikkerur for helpful feedback. This work was supported in part by C-BRIC (one of six centers in JUMP, a
Semiconductor Research Corporation (SRC) program sponsored by DARPA),
DARPA (HR00112190134) and the Army Research Office (W911NF2020053). The
authors affirm that the views expressed herein are solely their own, and
do not represent the views of the United States government or any agency
thereof.

\clearpage
\newpage
\bibliography{main_arxiv}

\begin{thebibliography}{44}
\providecommand{\natexlab}[1]{#1}
\providecommand{\url}[1]{\texttt{#1}}
\expandafter\ifx\csname urlstyle\endcsname\relax
  \providecommand{\doi}[1]{doi: #1}\else
  \providecommand{\doi}{doi: \begingroup \urlstyle{rm}\Url}\fi

\bibitem[Bitterwolf et~al.(2022)Bitterwolf, Meinke, Augustin, and
  Hein]{bitterwolfrevisiting}
Julian Bitterwolf, Alexander Meinke, Maximilian Augustin, and Matthias Hein.
\newblock Revisiting out-of-distribution detection: A simple baseline is
  surprisingly effective, 2022.
\newblock URL \url{https://openreview.net/forum?id=-BTmxCddppP}.

\bibitem[Chalapathy et~al.(2017)Chalapathy, Menon, and
  Chawla]{chalapathy2017robust}
Raghavendra Chalapathy, Aditya~Krishna Menon, and Sanjay Chawla.
\newblock Robust, deep and inductive anomaly detection.
\newblock In \emph{Machine Learning and Knowledge Discovery in Databases:
  European Conference, ECML PKDD 2017, Skopje, Macedonia, September 18--22,
  2017, Proceedings, Part I 10}, pages 36--51. Springer, 2017.

\bibitem[Chalapathy et~al.(2018)Chalapathy, Menon, and
  Chawla]{chalapathy2018anomaly}
Raghavendra Chalapathy, Aditya~Krishna Menon, and Sanjay Chawla.
\newblock Anomaly detection using one-class neural networks.
\newblock \emph{arXiv preprint arXiv:1802.06360}, 2018.

\bibitem[Choi et~al.(2018)Choi, Jang, and Alemi]{choi2018waic}
Hyunsun Choi, Eric Jang, and Alexander~A Alemi.
\newblock {WAIC}, but why? generative ensembles for robust anomaly detection.
\newblock \emph{arXiv preprint arXiv:1810.01392}, 2018.

\bibitem[Emmott et~al.(2015)Emmott, Das, Dietterich, Fern, and
  Wong]{emmott2015meta}
Andrew Emmott, Shubhomoy Das, Thomas Dietterich, Alan Fern, and Weng-Keen Wong.
\newblock A meta-analysis of the anomaly detection problem.
\newblock \emph{arXiv preprint arXiv:1503.01158}, 2015.

\bibitem[Esmaeilpour et~al.(2022)Esmaeilpour, Liu, Robertson, and
  Shu]{esmaeilpour2022zero}
Sepideh Esmaeilpour, Bing Liu, Eric Robertson, and Lei Shu.
\newblock Zero-shot out-of-distribution detection based on the pre-trained
  model {CLIP}.
\newblock In \emph{Proceedings of the AAAI conference on artificial
  intelligence}, pages 6568--6576, 2022.

\bibitem[Fort et~al.(2021)Fort, Ren, and Lakshminarayanan]{fort2021exploring}
Stanislav Fort, Jie Ren, and Balaji Lakshminarayanan.
\newblock Exploring the limits of out-of-distribution detection.
\newblock \emph{Advances in Neural Information Processing Systems},
  34:\penalty0 7068--7081, 2021.

\bibitem[Hendrycks and Gimpel(2016)]{hendrycks2016baseline}
Dan Hendrycks and Kevin Gimpel.
\newblock A baseline for detecting misclassified and out-of-distribution
  examples in neural networks.
\newblock \emph{arXiv preprint arXiv:1610.02136}, 2016.

\bibitem[Hendrycks et~al.(2018)Hendrycks, Mazeika, and
  Dietterich]{hendrycks2018deep}
Dan Hendrycks, Mantas Mazeika, and Thomas Dietterich.
\newblock Deep anomaly detection with outlier exposure.
\newblock \emph{arXiv preprint arXiv:1812.04606}, 2018.

\bibitem[Hendrycks et~al.(2019{\natexlab{a}})Hendrycks, Basart, Mazeika, Zou,
  Kwon, Mostajabi, Steinhardt, and Song]{hendrycks2019scaling}
Dan Hendrycks, Steven Basart, Mantas Mazeika, Andy Zou, Joe Kwon, Mohammadreza
  Mostajabi, Jacob Steinhardt, and Dawn Song.
\newblock Scaling out-of-distribution detection for real-world settings.
\newblock \emph{arXiv preprint arXiv:1911.11132}, 2019{\natexlab{a}}.

\bibitem[Hendrycks et~al.(2019{\natexlab{b}})Hendrycks, Zhao, Basart,
  Steinhardt, and Song]{hendrycks2019nae}
Dan Hendrycks, Kevin Zhao, Steven Basart, Jacob Steinhardt, and Dawn Song.
\newblock Natural adversarial examples.
\newblock \emph{arXiv preprint arXiv:1907.07174}, 2019{\natexlab{b}}.

\bibitem[Hendrycks et~al.(2020)Hendrycks, Basart, Mu, Kadavath, Wang, Dorundo,
  Desai, Zhu, Parajuli, Guo, Song, Steinhardt, and Gilmer]{hendrycks2020many}
Dan Hendrycks, Steven Basart, Norman Mu, Saurav Kadavath, Frank Wang, Evan
  Dorundo, Rahul Desai, Tyler Zhu, Samyak Parajuli, Mike Guo, Dawn Song, Jacob
  Steinhardt, and Justin Gilmer.
\newblock The many faces of robustness: A critical analysis of
  out-of-distribution generalization.
\newblock \emph{arXiv preprint arXiv:2006.16241}, 2020.

\bibitem[Hojjati and Armanfard(2021)]{hojjati2021dasvdd}
Hadi Hojjati and Narges Armanfard.
\newblock {DASVDD}: Deep autoencoding support vector data descriptor for
  anomaly detection.
\newblock \emph{arXiv preprint arXiv:2106.05410}, 2021.

\bibitem[Khan and Madden(2010)]{khan2010survey}
Shehroz~S Khan and Michael~G Madden.
\newblock A survey of recent trends in one class classification.
\newblock In \emph{Artificial Intelligence and Cognitive Science: 20th Irish
  Conference, AICS 2009, Dublin, Ireland, August 19-21, 2009, Revised Selected
  Papers 20}, pages 188--197. Springer, 2010.

\bibitem[Kirillov et~al.(2023)Kirillov, Mintun, Ravi, Mao, Rolland, Gustafson,
  Xiao, Whitehead, Berg, Lo, et~al.]{kirillov2023segment}
Alexander Kirillov, Eric Mintun, Nikhila Ravi, Hanzi Mao, Chloe Rolland, Laura
  Gustafson, Tete Xiao, Spencer Whitehead, Alexander~C Berg, Wan-Yen Lo, et~al.
\newblock Segment anything.
\newblock \emph{arXiv preprint arXiv:2304.02643}, 2023.

\bibitem[Lee et~al.(2018)Lee, Lee, Lee, and Shin]{lee2018simple}
Kimin Lee, Kibok Lee, Honglak Lee, and Jinwoo Shin.
\newblock A simple unified framework for detecting out-of-distribution samples
  and adversarial attacks.
\newblock \emph{NeurIPS}, 2018.

\bibitem[Li et~al.(2022)Li, Ling, Kim, Kreis, Fidler, and
  Torralba]{li2022bigdatasetgan}
Daiqing Li, Huan Ling, Seung~Wook Kim, Karsten Kreis, Sanja Fidler, and Antonio
  Torralba.
\newblock {BigDatasetGAN}: Synthesizing imagenet with pixel-wise annotations.
\newblock In \emph{Proceedings of the IEEE/CVF Conference on Computer Vision
  and Pattern Recognition}, pages 21330--21340, 2022.

\bibitem[Liang et~al.(2017)Liang, Li, and Srikant]{liang2017enhancing}
Shiyu Liang, Yixuan Li, and R~Srikant.
\newblock Enhancing the reliability of out-of-distribution image detection in
  neural networks.
\newblock \emph{arXiv preprint arXiv:1706.02690}, 2017.

\bibitem[Liu et~al.(2023)Liu, Zeng, Ren, Li, Zhang, Yang, Li, Yang, Su, Zhu,
  et~al.]{liu2023grounding}
Shilong Liu, Zhaoyang Zeng, Tianhe Ren, Feng Li, Hao Zhang, Jie Yang, Chunyuan
  Li, Jianwei Yang, Hang Su, Jun Zhu, et~al.
\newblock Grounding {DINO}: Marrying {DINO} with grounded pre-training for
  open-set object detection.
\newblock \emph{arXiv preprint arXiv:2303.05499}, 2023.

\bibitem[Liu et~al.(2020)Liu, Wang, Owens, and Li]{liu2020energy}
Weitang Liu, Xiaoyun Wang, John Owens, and Yixuan Li.
\newblock Energy-based out-of-distribution detection.
\newblock \emph{Advances in Neural Information Processing Systems},
  33:\penalty0 21464--21475, 2020.

\bibitem[Miller(1995)]{miller1995wordnet}
George~A Miller.
\newblock {WordNet}: a lexical database for {E}nglish.
\newblock \emph{Communications of the ACM}, 38\penalty0 (11):\penalty0 39--41,
  1995.

\bibitem[Ming et~al.(2022)Ming, Cai, Gu, Sun, Li, and Li]{ming2022delving}
Yifei Ming, Ziyang Cai, Jiuxiang Gu, Yiyou Sun, Wei Li, and Yixuan Li.
\newblock Delving into out-of-distribution detection with vision-language
  representations.
\newblock \emph{arXiv preprint arXiv:2211.13445}, 2022.

\bibitem[Morningstar et~al.(2021)Morningstar, Ham, Gallagher, Lakshminarayanan,
  Alemi, and Dillon]{morningstar2021density}
Warren Morningstar, Cusuh Ham, Andrew Gallagher, Balaji Lakshminarayanan, Alex
  Alemi, and Joshua Dillon.
\newblock Density of states estimation for out of distribution detection.
\newblock In \emph{International Conference on Artificial Intelligence and
  Statistics}, pages 3232--3240. PMLR, 2021.

\bibitem[Nalisnick et~al.(2018)Nalisnick, Matsukawa, Teh, Gorur, and
  Lakshminarayanan]{nalisnick2018deep}
Eric Nalisnick, Akihiro Matsukawa, Yee~Whye Teh, Dilan Gorur, and Balaji
  Lakshminarayanan.
\newblock Do deep generative models know what they don't know?
\newblock \emph{arXiv preprint arXiv:1810.09136}, 2018.

\bibitem[Noumir et~al.(2012)Noumir, Honeine, and Richard]{noumir2012simple}
Zineb Noumir, Paul Honeine, and Cedue Richard.
\newblock On simple one-class classification methods.
\newblock In \emph{2012 IEEE International Symposium on Information Theory
  Proceedings}, pages 2022--2026. IEEE, 2012.

\bibitem[Park et~al.(2021)Park, Moon, Ahn, and Sohn]{park2021wrong}
JuneKyu Park, Jeong-Hyeon Moon, Namhyuk Ahn, and Kyung-Ah Sohn.
\newblock What is wrong with one-class anomaly detection?
\newblock \emph{arXiv preprint arXiv:2104.09793}, 2021.

\bibitem[Radford et~al.(2021)Radford, Kim, Hallacy, Ramesh, Goh, Agarwal,
  Sastry, Askell, Mishkin, Clark, et~al.]{radford2021learning}
Alec Radford, Jong~Wook Kim, Chris Hallacy, Aditya Ramesh, Gabriel Goh,
  Sandhini Agarwal, Girish Sastry, Amanda Askell, Pamela Mishkin, Jack Clark,
  et~al.
\newblock Learning transferable visual models from natural language
  supervision.
\newblock In \emph{International conference on machine learning}, pages
  8748--8763. PMLR, 2021.

\bibitem[Recht et~al.(2019)Recht, Roelofs, Schmidt, and
  Shankar]{recht2019imagenet}
Benjamin Recht, Rebecca Roelofs, Ludwig Schmidt, and Vaishaal Shankar.
\newblock Do {ImageNet} classifiers generalize to imagenet?
\newblock In \emph{International Conference on Machine Learning}, pages
  5389--5400, 2019.

\bibitem[Ren et~al.(2019)Ren, Liu, Fertig, Snoek, Poplin, DePristo, Dillon, and
  Lakshminarayanan]{ren2019likelihood}
Jie Ren, Peter~J Liu, Emily Fertig, Jasper Snoek, Ryan Poplin, Mark~A DePristo,
  Joshua~V Dillon, and Balaji Lakshminarayanan.
\newblock Likelihood ratios for out-of-distribution detection.
\newblock \emph{NeurIPS}, 2019.

\bibitem[Ren et~al.(2021)Ren, Fort, Liu, Roy, Padhy, and
  Lakshminarayanan]{ren2021simple}
Jie Ren, Stanislav Fort, Jeremiah Liu, Abhijit~Guha Roy, Shreyas Padhy, and
  Balaji Lakshminarayanan.
\newblock A simple fix to {M}ahalanobis distance for improving near-ood
  detection.
\newblock \emph{arXiv preprint arXiv:2106.09022}, 2021.

\bibitem[Roy et~al.(2022)Roy, Ren, Azizi, Loh, Natarajan, Mustafa, Pawlowski,
  Freyberg, Liu, Beaver, et~al.]{roy2022does}
Abhijit~Guha Roy, Jie Ren, Shekoofeh Azizi, Aaron Loh, Vivek Natarajan, Basil
  Mustafa, Nick Pawlowski, Jan Freyberg, Yuan Liu, Zach Beaver, et~al.
\newblock Does your dermatology classifier know what it doesn’t know?
  {D}etecting the long-tail of unseen conditions.
\newblock \emph{Medical Image Analysis}, 75:\penalty0 102274, 2022.

\bibitem[Ruff et~al.(2018)Ruff, Vandermeulen, Goernitz, Deecke, Siddiqui,
  Binder, M{\"u}ller, and Kloft]{ruff2018deep}
Lukas Ruff, Robert Vandermeulen, Nico Goernitz, Lucas Deecke, Shoaib~Ahmed
  Siddiqui, Alexander Binder, Emmanuel M{\"u}ller, and Marius Kloft.
\newblock Deep one-class classification.
\newblock In \emph{International conference on machine learning}, pages
  4393--4402. PMLR, 2018.

\bibitem[Russakovsky et~al.(2015)Russakovsky, Deng, Su, Krause, Satheesh, Ma,
  Huang, Karpathy, Khosla, Bernstein, Berg, and Fei-Fei]{ILSVRC15}
Olga Russakovsky, Jia Deng, Hao Su, Jonathan Krause, Sanjeev Satheesh, Sean Ma,
  Zhiheng Huang, Andrej Karpathy, Aditya Khosla, Michael Bernstein,
  Alexander~C. Berg, and Li~Fei-Fei.
\newblock {ImageNet Large Scale Visual Recognition Challenge}.
\newblock \emph{International Journal of Computer Vision (IJCV)}, 115\penalty0
  (3):\penalty0 211--252, 2015.
\newblock \doi{10.1007/s11263-015-0816-y}.

\bibitem[Scholkopf et~al.(2000)Scholkopf, Williamson, Smola, Shawe-Taylor,
  Platt, et~al.]{scholkopf2000support}
Bernhard Scholkopf, Robert Williamson, Alex Smola, John Shawe-Taylor, John
  Platt, et~al.
\newblock Support vector method for novelty detection.
\newblock \emph{Advances in neural information processing systems}, 12\penalty0
  (3):\penalty0 582--588, 2000.

\bibitem[Shankar* et~al.(2020)Shankar*, Roelofs*, Mania, Fang, Recht, and
  Schmidt]{shankar2019evaluating}
Vaishaal Shankar*, Rebecca Roelofs*, Horia Mania, Alex Fang, Benjamin Recht,
  and Ludwig Schmidt.
\newblock Evaluating machine accuracy on {ImageNet}.
\newblock \emph{ICML}, 2020.

\bibitem[Sugiyama and Kawanabe(2012)]{sugiyama2012machine}
Masashi Sugiyama and Motoaki Kawanabe.
\newblock \emph{Machine learning in non-stationary environments: Introduction
  to covariate shift adaptation}.
\newblock MIT press, 2012.

\bibitem[Sun et~al.(2022)Sun, Ming, Zhu, and Li]{sun2022out}
Yiyou Sun, Yifei Ming, Xiaojin Zhu, and Yixuan Li.
\newblock Out-of-distribution detection with deep nearest neighbors.
\newblock \emph{arXiv preprint arXiv:2204.06507}, 2022.

\bibitem[Sundar et~al.(2020)Sundar, Ramakrishna, Rahiminasab, Easwaran, and
  Dubey]{sundar2020out}
Vijaya~Kumar Sundar, Shreyas Ramakrishna, Zahra Rahiminasab, Arvind Easwaran,
  and Abhishek Dubey.
\newblock Out-of-distribution detection in multi-label datasets using latent
  space of $\beta$-{VAE}.
\newblock In \emph{2020 IEEE Security and Privacy Workshops (SPW)}, pages
  250--255. IEEE, 2020.

\bibitem[Vasudevan et~al.(2022)Vasudevan, Caine, Gontijo-Lopes, Fridovich-Keil,
  and Roelofs]{vasudevan2022does}
Vijay Vasudevan, Benjamin Caine, Raphael Gontijo-Lopes, Sara Fridovich-Keil,
  and Rebecca Roelofs.
\newblock When does dough become a bagel? {A}nalyzing the remaining mistakes on
  {ImageNet}.
\newblock \emph{arXiv preprint arXiv:2205.04596}, 2022.

\bibitem[Wang et~al.(2019)Wang, Ge, Lipton, and Xing]{wang2019learning}
Haohan Wang, Songwei Ge, Zachary Lipton, and Eric~P Xing.
\newblock Learning robust global representations by penalizing local predictive
  power.
\newblock In \emph{Advances in Neural Information Processing Systems}, pages
  10506--10518, 2019.

\bibitem[Wang et~al.(2021)Wang, Liu, Bocchieri, and Li]{wangenergy}
Haoran Wang, Weitang Liu, Alex Bocchieri, and Yixuan Li.
\newblock Energy-based out-of-distribution detection for multi-label
  classification, 2021.
\newblock URL \url{https://openreview.net/forum?id=KsN9p5qJN3}.

\bibitem[Wang et~al.(2022)Wang, Huang, Huangfu, Liu, and Zhang]{wang2022multi}
Lei Wang, Sheng Huang, Luwen Huangfu, Bo~Liu, and Xiaohong Zhang.
\newblock Multi-label out-of-distribution detection via exploiting sparsity and
  co-occurrence of labels.
\newblock \emph{Image and Vision Computing}, 126:\penalty0 104548, 2022.

\bibitem[Zhai et~al.(2022)Zhai, Wang, Mustafa, Steiner, Keysers, Kolesnikov,
  and Beyer]{zhai2022lit}
Xiaohua Zhai, Xiao Wang, Basil Mustafa, Andreas Steiner, Daniel Keysers,
  Alexander Kolesnikov, and Lucas Beyer.
\newblock {LiT}: Zero-shot transfer with locked-image text tuning.
\newblock In \emph{Proceedings of the IEEE/CVF Conference on Computer Vision
  and Pattern Recognition}, pages 18123--18133, 2022.

\bibitem[Zhang and Deng(2021)]{zhang2021anomaly}
Zheng Zhang and Xiaogang Deng.
\newblock Anomaly detection using improved deep {SVDD} model with data
  structure preservation.
\newblock \emph{Pattern Recognition Letters}, 148:\penalty0 1--6, 2021.

\end{thebibliography}
\bibliographystyle{plainnat}

\newpage

\section*{Appendix}

% \section*{More results}

 Tables~\ref{tab:inet}-\ref{tab:inet-s} 
% Table~\ref{table1}, Table~\ref{tab:inet-v2}, Table~\ref{tab:inet-r}
% Table~\ref{tab:inet-a} and Table~\ref{tab:inet-s} 
%Table 1, Table 2, Table 3 Table 4 and Table 5
show additional one-class OOD detection results on ImageNet, ImageNet-v2, ImageNet-r, ImageNet-adversarial and ImageNet-Sketch respectively.

We use CLIP ViT-B/16 as the main model for evaluating our method in the main paper. %Table 6
Table~\ref{tab:inet-vitl}
shows the ablation study results on ImageNet with ViT-L/14 based CLIP model. Our scores outperformed the baselines on one-class OOD detection tasks, consistent with the main paper's conclusions.

%% Table 1 
\begin{table}[h]
\centering
\caption{Additional one-class OOD detection results on ImageNet}
\label{tab:inet}
\begin{tabular}{lll}
                    & $\mathcal{C}_{\texttt{in}}$ vs $\mathcal{C}_{\texttt{out}}$ &  $\mathcal{C}_{\texttt{in}}^\prime$ vs $\mathcal{C}_{\texttt{out}}^\prime$ \\ \toprule
% \multicolumn{3}{c}{Dog vs non-dog}                                                                                                                    \\
% $S_{\texttt{-max\_prob}}$&  0.7962    &   0.8647  \\
% $S_{\texttt{sum\_out\_prob}}$ &  0.9716     &   0.8250    \\
% $S_{\texttt{max\_out\_prob}}$ &   0.9972    &    0.7240 \\
% $S_{\texttt{-max\_in\_prob}}$ (ours) &   0.9983    &    0.9898 \\ 
% $S_{\texttt{max\_logit\_diff}}$ (ours) &   1.0000     &    0.9933 \\ 
% \midrule
% % $S_{\texttt{Ours + prompt aug}}$ &   1.0  &  0.998   &   0.999   &    0.982   \\ \midrule
% \multicolumn{3}{c}{Bird vs non-bird}        \\
% $S_{\texttt{-max\_prob}}$& 0.9767  & 0.8623 \\
% $S_{\texttt{sum\_out\_prob}}$ & 0.9814  & 0.8747 \\
% $S_{\texttt{max\_out\_prob}}$ & 0.9934  & 0.5746 \\
% $S_{\texttt{-max\_in\_prob}}$ (ours) & 0.9999  & 0.9943\\ 
% $S_{\texttt{max\_logit\_diff}}$ (ours) & 1.0000   & 0.9872\\ 
% \midrule    
% \multicolumn{3}{c}{Person vs non-person}                                                                                                                  \\
% $S_{\texttt{-max\_prob}}$& 0.9968  & 0.3288\\
% $S_{\texttt{sum\_out\_prob}}$ &0.7684  & 0.5696\\
% $S_{\texttt{max\_out\_prob}}$ &0.9920 & 0.5632 \\
% $S_{\texttt{-max\_in\_prob}}$ (ours) & 0.9999  & 0.6376 \\ 
% $S_{\texttt{max\_logit\_diff}}$ (ours) & 0.9998 & 0.6393\\ 
% \midrule    
\multicolumn{3}{c}{Bird vs non-bird}                                                                                                                  \\
$S_{\texttt{-max\_prob}}$& 0.9827  & 0.8609 \\
$S_{\texttt{sum\_out\_prob}}$ & 0.9781 & 0.8067 \\
$S_{\texttt{max\_out\_prob}}$ & 0.9943 & 0.5443 \\
$S_{\texttt{-max\_in\_prob}}$ (ours) & \underline{0.9999}  &\textbf{0.9912} \\ 
$S_{\texttt{max\_logit\_diff}}$ (ours) & \textbf{1.0000}  & \underline{0.9810}\\ 
\midrule    
\multicolumn{3}{c}{Chair vs non-chair}                                                                                                                  \\
$S_{\texttt{-max\_prob}}$& 0.7159  & 0.3198 \\
$S_{\texttt{sum\_out\_prob}}$ & 0.9616  &  0.6338 \\
$S_{\texttt{max\_out\_prob}}$ &  0.9910  & 0.7094 \\
$S_{\texttt{-max\_in\_prob}}$ (ours) & \underline{0.9967}  & \textbf{0.9820} \\ 
$S_{\texttt{max\_logit\_diff}}$ (ours) & \textbf{0.9999}  & \underline{0.9785} \\ 
\midrule    
\multicolumn{3}{c}{Cat vs non-cat}                                                                                                                  \\
$S_{\texttt{-max\_prob}}$& 0.7456   &  0.6732 \\
$S_{\texttt{sum\_out\_prob}}$ & 0.6262  &  0.5324 \\
$S_{\texttt{max\_out\_prob}}$ &  0.9843  & 0.5554\\
$S_{\texttt{-max\_in\_prob}}$ (ours) & \underline{0.9994} & \textbf{0.9952} \\ 
$S_{\texttt{max\_logit\_diff}}$ (ours) & \textbf{1.0000}  & \underline{0.9916} \\ 
% \midrule    
% \multicolumn{5}{c}{Bus vs non-bus}                                                                                                                  \\
% $S_{\texttt{sum\_out\_prob}}$ & 0.6816 & 0.3242 & 0.6486 & 0.2572  \\
% $S_{\texttt{-max\_prob}}$& 0.8699 & 0.8883 & 0.5700 & 0.6301 \\
% $S_{\texttt{max\_out\_prob}}$ & 0.9681 & 0.3210 & 0.9890 & 0.2861 \\
% $S_{\texttt{-max\_in\_prob}}$ & 0.9989 & 0.9994 & 0.9961 & 0.9982\\ 
% $S_{\texttt{max\_logit\_diff}}$ (ours) &0.9998 & 0.9982 &  0.9997 & 0.9977\\ 
\bottomrule    
\end{tabular}
\end{table}

\begin{table}[h]
\centering
\caption{Additional one-class OOD detection results on ImageNet-v2}
\label{tab:inet-v2}
\begin{tabular}{lll}
                    & $\mathcal{C}_{\texttt{in}}$ vs $\mathcal{C}_{\texttt{out}}$ & $\mathcal{C}_{\texttt{in}}^\prime$ vs $\mathcal{C}_{\texttt{out}}^\prime$ \\ \toprule
% \multicolumn{3}{c}{Dog vs non-dog}                                                                                                                    \\
% $S_{\texttt{-max\_prob}}$& 0.7163 & 0.7384 \\
% $S_{\texttt{sum\_out\_prob}}$ & 0.9590  &  0.8083  \\
% $S_{\texttt{max\_out\_prob}}$ & 0.9923 & 0.7067 \\
% $S_{\texttt{-max\_in\_prob}}$ (ours) & 0.9945  & 0.9795\\ 
% $S_{\texttt{max\_logit\_diff}}$ (ours) & 0.9994  &  0.9836 \\ 
% \midrule
% % $S_{\texttt{Ours + prompt aug}}$ &   1.0  &  0.998   &   0.999   &    0.982   \\ \midrule
% \multicolumn{3}{c}{Bird vs non-bird}        \\
% $S_{\texttt{-max\_prob}}$& 0.9584 & 0.7855 \\
% $S_{\texttt{sum\_out\_prob}}$ & 0.9770 &  0.8190 \\
% $S_{\texttt{max\_out\_prob}}$ &0.9859 & 0.5345 \\
% $S_{\texttt{-max\_in\_prob}}$ (ours) & 0.9989 & 0.9850\\ 
% $S_{\texttt{max\_logit\_diff}}$ (ours) & 1.0000 & 0.9745\\ 
% \midrule    
% \multicolumn{3}{c}{Person vs non-person}                                                                                                                  \\
% $S_{\texttt{-max\_prob}}$& 0.5000  &  0.5000 \\
% $S_{\texttt{sum\_out\_prob}}$ &0.7144  &  0.4766  \\
% $S_{\texttt{max\_out\_prob}}$ & 0.9960  & 0.6050 \\
% $S_{\texttt{-max\_in\_prob}}$ (ours) & 0.9893 & 0.5519\\ 
% $S_{\texttt{max\_logit\_diff}}$ (ours) & 0.9999 & 0.5915 \\ 
% \midrule    
\multicolumn{3}{c}{Bird vs non-bird}                                                                                                                  \\
$S_{\texttt{-max\_prob}}$& 0.9770  & 0.8190  \\
$S_{\texttt{sum\_out\_prob}}$ & 0.9584  & 0.7855  \\
$S_{\texttt{max\_out\_prob}}$ &  0.9859 &  0.5345 \\
$S_{\texttt{-max\_in\_prob}}$ (ours) & \underline{0.9989}  &  \textbf{0.9850}\\ 
$S_{\texttt{max\_logit\_diff}}$ (ours) & \textbf{1.0000}  & \underline{0.9745} \\ 
\midrule    
\multicolumn{3}{c}{Chair vs non-chair}                                                                                                                  \\
$S_{\texttt{-max\_prob}}$& 0.8098  & 0.5739  \\
$S_{\texttt{sum\_out\_prob}}$ & 0.6639  &  0.4310  \\
$S_{\texttt{max\_out\_prob}}$ & 0.9630 & 0.6300 \\
$S_{\texttt{-max\_in\_prob}}$ (ours) & \underline{0.9931}  &  \textbf{0.9774} \\ 
$S_{\texttt{max\_logit\_diff}}$ (ours) & \textbf{0.9978}  & \underline{0.9552}\\ 
\midrule    
\multicolumn{3}{c}{Cat vs non-cat}                                                                                                                  \\
$S_{\texttt{-max\_prob}}$& 0.6488   &   0.5521  \\
$S_{\texttt{sum\_out\_prob}}$ & 0.8052   &  0.5865 \\
$S_{\texttt{max\_out\_prob}}$ & 0.9909   & 0.5841 \\
$S_{\texttt{-max\_in\_prob}}$ (ours) & \underline{0.9988}   & \textbf{0.9937} \\ 
$S_{\texttt{max\_logit\_diff}}$ (ours) & \textbf{1.0000} & \underline{0.9683}\\ 
% \midrule    
% \multicolumn{5}{c}{Bus vs non-bus}                                                                                                                  \\
% $S_{\texttt{sum\_out\_prob}}$ &  0.7180 &  0.3650  &    0.6656  &  0.2912     \\
% $S_{\texttt{-max\_prob}}$&   0.5000  &  0.5000   & 0.5000  & 0.5000    \\
% $S_{\texttt{max\_out\_prob}}$ &  0.9752  &  0.3383  &  0.9618   & 0.3486   \\
% $S_{\texttt{-max\_in\_prob}}$ &  0.9999  & 0.9994   &   0.9984   &  0.9953   \\ 
% $S_{\texttt{max\_logit\_diff}}$ (ours) &  1.0000   &  0.9995   &  0.9982    &   0.9672  \\ 
\bottomrule    
\end{tabular}
\end{table}

%%% Table 3 ImageNet-r
\begin{table}[h]
\centering
\caption{Additional one-class OOD detection results on ImageNet-r}
\label{tab:inet-r}
\begin{tabular}{lll}
                    & $\mathcal{C}_{\texttt{in}}$ vs $\mathcal{C}_{\texttt{out}}$  & $\mathcal{C}_{\texttt{in}}^\prime$ vs $\mathcal{C}_{\texttt{out}}^\prime$ \\ \toprule
% \multicolumn{3}{c}{Dog vs non-dog}                                                                                                                    \\
% $S_{\texttt{-max\_prob}}$&  0.8148     &  0.6334   \\
% $S_{\texttt{sum\_out\_prob}}$ &  0.9616     &  0.7812     \\
% $S_{\texttt{max\_out\_prob}}$ & 0.9758    &  0.6950   \\
% $S_{\texttt{-max\_in\_prob}}$ (ours) &  0.9903    &  0.9733   \\ 
% $S_{\texttt{max\_logit\_diff}}$ (ours) &  0.9990      &   0.9726  \\ 
% \midrule
% % $S_{\texttt{Ours + prompt aug}}$ &   1.0  &  0.998   &   0.999   &    0.982   \\ \midrule
% \multicolumn{3}{c}{Bird vs non-bird}        \\
% $S_{\texttt{-max\_prob}}$&  0.8286     &  0.5350   \\
% $S_{\texttt{sum\_out\_prob}}$ &  0.9346     &    0.6155   \\
% $S_{\texttt{max\_out\_prob}}$ &  0.9820    &  0.6432   \\
% $S_{\texttt{-max\_in\_prob}}$ (ours) & 0.9874    &   0.9528  \\ 
% $S_{\texttt{max\_logit\_diff}}$ (ours) &  0.9993     &  0.9477   \\ 
% \midrule    
% \multicolumn{3}{c}{Person vs non-person}                                                                                                                  \\
% $S_{\texttt{-max\_prob}}$&  0.5000     &   0.5000  \\
% $S_{\texttt{sum\_out\_prob}}$ & 0.8639      &   0.5818    \\
% $S_{\texttt{max\_out\_prob}}$ & 0.9662     &  0.5917   \\
% $S_{\texttt{-max\_in\_prob}}$ (ours) &  0.9994      &  0.6360   \\ 
% $S_{\texttt{max\_logit\_diff}}$ (ours) &  0.9995      &   0.6559  \\ 
% \midrule    
\multicolumn{3}{c}{Bird vs non-bird}                                                                                                                  \\
$S_{\texttt{-max\_prob}}$&   0.9346    &  0.6155   \\
$S_{\texttt{sum\_out\_prob}}$ &  0.8286     &  0.5350     \\
$S_{\texttt{max\_out\_prob}}$ &  0.9820    &   0.6432  \\
$S_{\texttt{-max\_in\_prob}}$ (ours) & \underline{0.9874}     &  \textbf{0.9528}   \\ 
$S_{\texttt{max\_logit\_diff}}$ (ours) & \textbf{0.9993}    &  \underline{0.9477}   \\ 
\bottomrule    
\end{tabular}
\end{table}

%%% Table 4 ImageNet-adversarial
\begin{table}[h]
\centering
\caption{Additional one-class OOD detection results on ImageNet-adversarial}
\label{tab:inet-a}
\begin{tabular}{lll}
                    & $\mathcal{C}_{\texttt{in}}$ vs $\mathcal{C}_{\texttt{out}}$ & $\mathcal{C}_{\texttt{in}}^\prime$ vs $\mathcal{C}_{\texttt{out}}^\prime$ \\ \toprule
% \multicolumn{3}{c}{Dog vs non-dog}                                                                                                                    \\
% $S_{\texttt{-max\_prob}}$&  0.3544     &  0.3991   \\
% $S_{\texttt{sum\_out\_prob}}$ &  0.9399    &    0.6848   \\
% $S_{\texttt{max\_out\_prob}}$ &  0.9307    &  0.6778   \\
% $S_{\texttt{-max\_in\_prob}}$ (ours) &  0.8963     &   0.9261  \\ 
% $S_{\texttt{max\_logit\_diff}}$ (ours) &  0.9769    &  0.9333   \\ 
% \midrule
% % $S_{\texttt{Ours + prompt aug}}$ &   1.0  &  0.998   &   0.999   &    0.982   \\ \midrule
% \multicolumn{3}{c}{Bird vs non-bird}        \\
% $S_{\texttt{-max\_prob}}$&  0.7546    &  0.5489   \\
% $S_{\texttt{sum\_out\_prob}}$ &  0.8647     &    0.5002   \\
% $S_{\texttt{max\_out\_prob}}$ &  0.8896   &  0.6015   \\
% $S_{\texttt{-max\_in\_prob}}$ (ours) & 0.9660    &  0.9212   \\ 
% $S_{\texttt{max\_logit\_diff}}$ (ours) &  0.9870     &   0.9238  \\ 
% \midrule    
\multicolumn{3}{c}{Bird vs non-bird}                                                                                                                  \\
$S_{\texttt{-max\_prob}}$&   0.8647   &  0.5002   \\
$S_{\texttt{sum\_out\_prob}}$ &  0.7546    &   0.5489    \\
$S_{\texttt{max\_out\_prob}}$ &   0.8896   &   0.6015  \\
$S_{\texttt{-max\_in\_prob}}$ (ours) & \underline{0.9660}     &  \underline{0.9212}   \\ 
$S_{\texttt{max\_logit\_diff}}$ (ours) &  \textbf{0.9870}      &  \textbf{0.9238}   \\ 
\midrule    
\multicolumn{3}{c}{Cat vs non-cat}                                                                                                                  \\
$S_{\texttt{-max\_prob}}$&  0.4583    &  0.6136   \\
$S_{\texttt{sum\_out\_prob}}$ &  0.5008     &   0.1434    \\
$S_{\texttt{max\_out\_prob}}$ &  0.8315    &  0.3491   \\
$S_{\texttt{-max\_in\_prob}}$ (ours) &  \textbf{0.9951}    &  \textbf{0.9609}   \\ 
$S_{\texttt{max\_logit\_diff}}$ (ours) &  \underline{0.9878}     &  \underline{0.9009}   \\ 
\bottomrule    
\end{tabular}
\end{table}

%%% Table 5 ImageNet-Sketch
\begin{table}[h]
\centering
\caption{Additional one-class OOD detection results on ImageNet-Sketch}
\label{tab:inet-s}
\begin{tabular}{lll}
                    & $\mathcal{C}_{\texttt{in}}$ vs $\mathcal{C}_{\texttt{out}}$ & $\mathcal{C}_{\texttt{in}}^\prime$ vs $\mathcal{C}_{\texttt{out}}^\prime$ \\ \toprule
% \multicolumn{3}{c}{Dog vs non-dog}                                                                                                                    \\
% $S_{\texttt{-max\_prob}}$&  0.7676    &   0.8073  \\
% $S_{\texttt{sum\_out\_prob}}$ & 0.9643      & 0.8232\\
% $S_{\texttt{max\_out\_prob}}$ &  0.9820    &  0.6979   \\
% $S_{\texttt{-max\_in\_prob}}$ (ours) & 0.9851    &  0.9868   \\ 
% $S_{\texttt{max\_logit\_diff}}$ (ours) &  0.9993     &   0.9850  \\ 
% \midrule
% % $S_{\texttt{Ours + prompt aug}}$ &   1.0  &  0.998   &   0.999   &    0.982   \\ \midrule
% \multicolumn{3}{c}{Bird vs non-bird}        \\
% $S_{\texttt{-max\_prob}}$&  0.8430     &   0.7789  \\
% $S_{\texttt{sum\_out\_prob}}$ &  0.9669    &   0.6744    \\
% $S_{\texttt{max\_out\_prob}}$ & 0.9760     &   0.5300  \\
% $S_{\texttt{-max\_in\_prob}}$ (ours) & 0.9900      &  0.9860   \\ 
% $S_{\texttt{max\_logit\_diff}}$ (ours) &  0.9996      &  0.9845   \\ 
% \midrule    
% \multicolumn{3}{c}{Person vs non-person}                                                                                                                  \\
% $S_{\texttt{-max\_prob}}$&  0.5000     &  0.5000   \\
% $S_{\texttt{sum\_out\_prob}}$ &  0.7984       &   0.4351    \\
% $S_{\texttt{max\_out\_prob}}$ & 0.9861  & 0.6698 \\
% $S_{\texttt{-max\_in\_prob}}$ (ours) & 1.0000  & 0.8347 \\ 
% $S_{\texttt{max\_logit\_diff}}$ (ours) & 1.0000  & 0.8254 \\ 
% \midrule
\multicolumn{3}{c}{Bird vs non-bird}                                                                                                                  \\
$S_{\texttt{-max\_prob}}$&  0.9669    &   0.6744  \\
$S_{\texttt{sum\_out\_prob}}$ &   0.8430    &   0.7789    \\
$S_{\texttt{max\_out\_prob}}$ &  0.9760    &   0.5300  \\
$S_{\texttt{-max\_in\_prob}}$ (ours) &  \underline{0.9900}     & \textbf{0.9860}    \\ 
$S_{\texttt{max\_logit\_diff}}$ (ours) &  \textbf{0.9996}      &  \underline{0.9845}   \\ 
\midrule    
\multicolumn{3}{c}{Chair vs non-chair}                                                                                                                  \\
$S_{\texttt{-max\_prob}}$&   0.8841   &  0.7309   \\
$S_{\texttt{sum\_out\_prob}}$ &  0.8091     &  0.3212     \\
$S_{\texttt{max\_out\_prob}}$ &  0.9826     &  0.6285   \\
$S_{\texttt{-max\_in\_prob}}$ (ours) & \underline{0.9997}     &  \textbf{0.9912}   \\ 
$S_{\texttt{max\_logit\_diff}}$ (ours) &  \textbf{1.0000}    &  \underline{0.9892}   \\ 
\midrule    
\multicolumn{3}{c}{Cat vs non-cat}                                                                                                                  \\
$S_{\texttt{-max\_prob}}$&  0.5086     &  0.3775   \\
$S_{\texttt{sum\_out\_prob}}$ & 0.7868       &   0.3201    \\
$S_{\texttt{max\_out\_prob}}$ &0.9571  &  0.4238   \\
$S_{\texttt{-max\_in\_prob}}$ (ours) &  \underline{0.9931}     &   \textbf{0.9915}  \\ 
$S_{\texttt{max\_logit\_diff}}$ (ours) &  \textbf{0.9991}     &  \underline{0.9787}   \\  
% \midrule
% \multicolumn{5}{c}{Bus vs non-bus}                                                                                                                  \\
% $S_{\texttt{sum\_ood(softmax(in, out))}}$ & 0.8475   &  0.4026    &   0.8270   &   0.3604    \\
% $S_{\texttt{-max\_prob}}$&  0.5000   &  0.5000   & 0.5000  &  0.5000   \\
% $S_{\texttt{max\_out\_prob}}$ &  0.9313  &  0.1701  &  0.8540   &  0.1175   \\
% $S_{\texttt{-max\_in\_prob}}$ &  0.9999  &  0.9993  &  0.9920    &   0.9774  \\ 
% $S_{\texttt{max\_logit\_diff}}$ (ours) &  0.9999   &  0.9990   &  0.9742    &  0.8230   \\ 
\bottomrule    
\end{tabular}
\end{table}

%%% Table 6 ViT L-14 result
\begin{table}[h]
\centering
\caption{One-class OOD detection results on ImageNet with ViT-L/14 based CLIP model}
\label{tab:inet-vitl}
\begin{tabular}{lll}
                    & $\mathcal{C}_{\texttt{in}}$ vs $\mathcal{C}_{\texttt{out}}$ & $\mathcal{C}_{\texttt{in}}^\prime$ vs $\mathcal{C}_{\texttt{out}}^\prime$ \\ \toprule
\multicolumn{3}{c}{Dog vs non-dog}                                                                                                                  \\
$S_{\texttt{-max\_prob}}$&   0.9852    &   0.8802   \\
$S_{\texttt{sum\_out\_prob}}$ &   0.8684    &   0.7201    \\
$S_{\texttt{max\_out\_prob}}$ &   0.9966   &   0.6640   \\
$S_{\texttt{-max\_in\_prob}}$ (ours) &   \underline{0.9986}   &  \underline{0.9578}  \\ 
$S_{\texttt{max\_logit\_diff}}$ (ours) &    \textbf{1.0000}     &   \textbf{0.9591}   \\ 
\midrule 
\multicolumn{3}{c}{Car vs non-car}                                                                                                                  \\
$S_{\texttt{-max\_prob}}$&   0.8240    &  0.5324   \\
$S_{\texttt{sum\_out\_prob}}$ &   0.8368     &  0.6152     \\
$S_{\texttt{max\_out\_prob}}$ &    0.9688   &  0.4115   \\
$S_{\texttt{-max\_in\_prob}}$ (ours) &   \underline{0.9886}   &  \textbf{0.9699}  \\ 
$S_{\texttt{max\_logit\_diff}}$ (ours) &    \textbf{0.9989}     &  \underline{0.9136}    \\ 
\midrule  

\multicolumn{3}{c}{Person vs non-person}                                                                                                                  \\
$S_{\texttt{-max\_prob}}$&    0.7701   &   0.5207   \\
$S_{\texttt{sum\_out\_prob}}$ &    0.9873    &   \textbf{0.7661}   \\
$S_{\texttt{max\_out\_prob}}$ &   0.9901    &   \underline{0.7458}   \\
$S_{\texttt{-max\_in\_prob}}$ (ours) &   \underline{0.9996}   &  0.5790  \\ 
$S_{\texttt{max\_logit\_diff}}$ (ours) &    \textbf{1.0000}     &   0.7131   \\ 
\midrule 
  
\multicolumn{3}{c}{Bird vs non-bird}                                                                                                                  \\
$S_{\texttt{-max\_prob}}$&   0.9825    &   0.8784   \\
$S_{\texttt{sum\_out\_prob}}$ &   0.9660     &   0.6642    \\
$S_{\texttt{max\_out\_prob}}$ &   0.9944    &  0.5468    \\
$S_{\texttt{-max\_in\_prob}}$ (ours) &   \underline{0.9998}   &  \textbf{0.9644}  \\ 
$S_{\texttt{max\_logit\_diff}}$ (ours) &    \textbf{1.0000}     &  \underline{0.9501}    \\ 
\midrule 
\multicolumn{3}{c}{Chair vs non-chair}                                                                                                                  \\
$S_{\texttt{-max\_prob}}$&    0.6482   &   0.3487  \\
$S_{\texttt{sum\_out\_prob}}$ &    0.9611    &   0.6423    \\
$S_{\texttt{max\_out\_prob}}$ &   0.9878    &  0.6630    \\
$S_{\texttt{-max\_in\_prob}}$ (ours) &  \underline{0.9969}    &  \textbf{0.9874}  \\ 
$S_{\texttt{max\_logit\_diff}}$ (ours) &   \textbf{1.0000}      &   \underline{0.9828}   \\ 
\midrule    
\multicolumn{3}{c}{Cat vs non-cat}                                                                                                                  \\
$S_{\texttt{-max\_prob}}$&   0.8027    &   0.6915   \\
$S_{\texttt{sum\_out\_prob}}$ &    0.6384    &   0.4935   \\
$S_{\texttt{max\_out\_prob}}$ &   0.9928    &  0.5705    \\
$S_{\texttt{-max\_in\_prob}}$ (ours) &    \underline{0.9998}  & \textbf{0.9934}   \\ 
$S_{\texttt{max\_logit\_diff}}$ (ours) &    \textbf{1.0000}     &  \underline{0.9892}    \\ 
% \multicolumn{3}{c}{Bus vs non-bus}                                                                                                                  \\
% $S_{\texttt{-max\_prob}}$&       &     \\
% $S_{\texttt{sum\_out\_prob}}$ &       &       \\
% $S_{\texttt{max\_out\_prob}}$ &      &      \\
% $S_{\texttt{-max\_in\_prob}}$ (ours) &  \underline{}    &  \textbf{}  \\ 
% $S_{\texttt{max\_logit\_diff}}$ (ours) &   \textbf{}      &   \underline{}   \\ 
% % \midrule 
% \multicolumn{5}{c}{Bus vs non-bus}                                                                                                                  \\
% $S_{\texttt{sum\_ood(softmax(in, out))}}$ & 0.8475   &  0.4026    &   0.8270   &   0.3604    \\
% $S_{\texttt{-max\_prob}}$&  0.5000   &  0.5000   & 0.5000  &  0.5000   \\
% $S_{\texttt{max\_out\_prob}}$ &  0.9313  &  0.1701  &  0.8540   &  0.1175   \\
% $S_{\texttt{-max\_in\_prob}}$ &  0.9999  &  0.9993  &  0.9920    &   0.9774  \\ 
% $S_{\texttt{max\_logit\_diff}}$ (ours) &  0.9999   &  0.9990   &  0.9742    &  0.8230   \\ 
\bottomrule    
\end{tabular}
\end{table}

\end{document}